\documentclass[lettersize,journal]{IEEEtran}
\usepackage{amsmath,amsfonts}
\usepackage{algorithmic}
\usepackage{algorithm}
\usepackage{array}
\usepackage[caption=false,font=normalsize,labelfont=sf,textfont=sf]{subfig}
\usepackage{textcomp}
\usepackage{stfloats}
\usepackage{url}
\usepackage{verbatim}
\usepackage{graphicx}
\usepackage{cite}
\usepackage{color}
\usepackage{booktabs}
\usepackage{amsfonts}
\usepackage{multirow}
\usepackage{adjustbox}
\usepackage{threeparttable}
\begin{document}

\title{SEA++: Multi-Graph-based High-Order Sensor Alignment for Multivariate Time-Series Unsupervised Domain Adaptation}
\author{Yucheng Wang, Yuecong Xu, Jianfei Yang, Min Wu, Xiaoli Li, Lihua Xie, \textit{Fellow, IEEE}, Zhenghua Chen\footnote{Corresponding Author}
\thanks{Yucheng Wang and Yuecong Xu are with Institute for Infocomm Research, A$^*$STAR, Singapore and the School of Electrical and Electronic Engineering, Nanyang Technological University, Singapore (Email: yucheng003@e.ntu.edu.sg, xuyu0014@e.ntu.edu.sg).}
\thanks{Zhenghua Chen and Min Wu are with Institute for Infocomm Research, A$^*$STAR, Singapore (Email: chen0832@e.ntu.edu.sg, wumin@i2r.a-star.edu.sg).}
\thanks{Xiaoli Li is with Institute for Infocomm Research, A$^*$STAR, Singapore and the School of Computer Science and Engineering, Nanyang Technological University, Singapore (Email: xlli@i2r.a-star.edu.sg).}
\thanks{Jianfei Yang and Lihua Xie are with the School of Electrical and Electronic Engineering, Nanyang Technological University, Singapore (Email: yang0478@e.ntu.edu.sg, elhxie@ntu.edu.sg).}
}

\markboth{Journal of \LaTeX\ Class Files,~Vol.~14, No.~8, August~2021}%
{Shell \MakeLowercase{\textit{et al.}}: A Sample Article Using IEEEtran.cls for IEEE Journals}


\maketitle

\begin{abstract}
Unsupervised Domain Adaptation (UDA) methods have been successful in reducing label dependency by minimizing the domain discrepancy between a labeled source domain and an unlabeled target domain. However, these methods face challenges when dealing with Multivariate Time-Series (MTS) data. MTS data typically consist of multiple sensors, each with its own unique distribution. This characteristic makes it hard to adapt existing UDA methods, which mainly focus on aligning global features while overlooking the distribution discrepancies at the sensor level, to reduce domain discrepancies for MTS data. To address this issue, a practical domain adaptation scenario is formulated as Multivariate Time-Series Unsupervised Domain Adaptation (MTS-UDA). In this paper, we propose SEnsor Alignment (SEA) for MTS-UDA, aiming to reduce domain discrepancy at both the local and global sensor levels. At the local sensor level, we design endo-feature alignment, which aligns sensor features and their correlations across domains. To reduce domain discrepancy at the global sensor level, we design exo-feature alignment that enforces restrictions on global sensor features. {\color{black}We further extend SEA to SEA++ by enhancing the endo-feature alignment. Particularly, we incorporate multi-graph-based high-order alignment for both sensor features and their correlations. High-order statistics are employed to achieve comprehensive alignment by capturing complex data distributions across domains. Meanwhile, a multi-graph alignment technique is introduced to effectively align the evolving distributions of MTS data.} Extensive empirical results have demonstrated the state-of-the-art performance of our SEA and SEA++ on public MTS datasets for MTS-UDA.

Unsupervised Domain Adaptation (UDA) methods have been successful in reducing label dependency by minimizing the domain discrepancy between a labeled source domain and an unlabeled target domain. However, these methods face challenges when dealing with Multivariate Time-Series (MTS) data. MTS data typically consist of multiple sensors, each with its own unique distribution. This characteristic makes it hard to adapt existing UDA methods, which mainly focus on aligning global features while overlooking the distribution discrepancies at the sensor level, to reduce domain discrepancies for MTS data. To address this issue, a practical domain adaptation scenario is formulated as Multivariate Time-Series Unsupervised Domain Adaptation (MTS-UDA). In this paper, we propose SEnsor Alignment (SEA) for MTS-UDA, aiming to reduce domain discrepancy at both the local and global sensor levels. At the local sensor level, we design endo-feature alignment, which aligns sensor features and their correlations across domains. To reduce domain discrepancy at the global sensor level, we design exo-feature alignment that enforces restrictions on global sensor features. We further extend SEA to SEA++ by enhancing the endo-feature alignment. Particularly, we incorporate multi-graph-based high-order alignment for both sensor features and their correlations. Extensive empirical results have demonstrated the state-of-the-art performance of our SEA and SEA++ on public MTS datasets for MTS-UDA. 

\end{abstract}



\section{Introduction}
\IEEEPARstart{T}ime-Series (TS) data have gained significant attention in various fields due to their wide range of applications. Deep Learning (DL) methods have played a pivotal role in addressing TS-related problems by leveraging their ability to effectively capture latent dependencies within the data \cite{eldele2023self,ma2020adversarial,zhao2019deep, chen2021deep, He_Zhang_Bai_Yi_Niu_2022,zheng2022multivariate,garg2021evaluation}. However, the effectiveness of DL models heavily relies on the availability of large-scale labeled TS data, which is often a challenge in real-world applications due to the high cost associated with data labeling.

To reduce the cost of labeling, Unsupervised Domain Adaptation (UDA) methods have been proposed to transfer the knowledge from a labeled source domain to an unlabeled target domain, enabling effective learning in the absence of target domain labels \cite{zhao2020review}. Due to domain shifts, existing UDA methods focused on reducing discrepancy across domains by learning domain-invariant features \cite{https://doi.org/10.48550/arxiv.1412.3474,10.5555/2946645.2946704,8099799,long2018conditional,10.1007/978-3-319-49409-8_35,long2017deep}.
These methods have achieved decent performance, showing the effectiveness of UDA in reducing label dependency. Subsequently, researchers applied UDA to TS data, aiming to reduce domain discrepancy by learning domain-invariant temporal features extracted by Recurrent Neural Network (RNN) \cite{Purushotham2017VariationalRA}, Long Short-Term Memory (LSTM) \cite{Cai_Chen_Li_Chen_Zhang_Ye_Li_Yang_Zhang_2021}, and Convolutional Neural Network (CNN) \cite{ijcai2021-378}.

\begin{figure}[!t]
    \centering
    \includegraphics[width = 1.\linewidth]{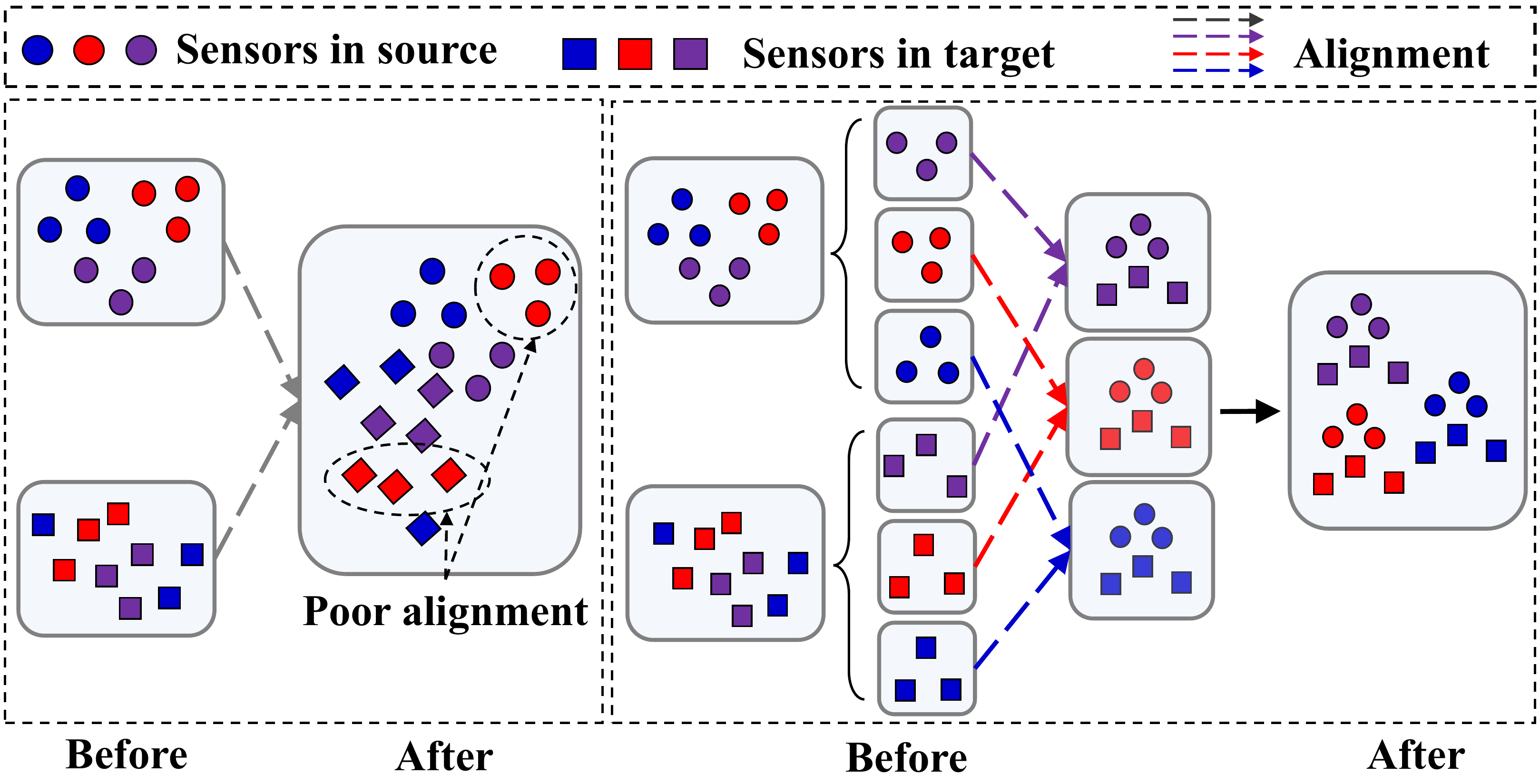}
    \caption{Comparisons between existing UDA methods and ours, where before and after represent before and after alignment respectively.
    {Left}: Only global sensor features are aligned, resulting in the poor alignment of the red sensor. {Right}: SEA aligns the sensor information between domains, so the feature discrepancy of each sensor can be reduced.}
    \vspace{-0.1cm}
    \label{fig:SEAnece}
\end{figure}

While substantial progress has been made, existing TS UDA methods may be inapplicable for real-world applications where multiple sensors are deployed simultaneously. For instance, in Remaining Useful Life (RUL) prediction, various types of sensors are employed to measure different physical parameters. On the other hand, in Human Activity Recognition (HAR), multiple sensors are positioned at various locations on the human body to capture activity data. To address these application scenarios, we formulate a more challenging yet practical problem as Multivariate Time-Series Unsupervised Domain Adaptation (MTS-UDA). Applying current TS UDA methods directly to MTS-UDA poses two key limitations. First, MTS data typically consist of signals from multiple sensors, where each sensor collects data with distinct distributions due to variations in sensor types or deployment locations \cite{shi2022deep}. When adapting existing TS UDA methods to MTS-UDA, they mainly focus on learning global features to facilitate alignment across domains by treating the signals from various sensors as a whole \cite{Cai_Chen_Li_Chen_Zhang_Ye_Li_Yang_Zhang_2021,https://doi.org/10.48550/arxiv.2203.08321}. However, these methods ignore the diverse distributions across sensors, resulting in misalignment at the sensor level and thus leading to suboptimal solutions for MTS-UDA. As presented in Fig.~\ref{fig:SEAnece} (\textbf{Left}), existing UDA methods focused on aligning the global features between domains, resulting in the poor alignment of the features from the red sensor and limiting the transferability of the model. 
Second, MTS data contain essential spatial-temporal dependency information due to its multi-source nature. Specifically, the spatial dependency refers to the correlations between sensors, which capture the critical interactive relationships among them. For instance, the spatial dependency in the context of machine RUL prediction corresponds to the correlation between a temperature sensor and a fan speed sensor, as they exhibit strong correlations. Besides, the temporal dependency refers to the temporal relationships between consecutive timestamps. Existing UDA methods are limited in effectively modelling and transferring both dependencies across domains, thus leading to suboptimal solutions for MTS-UDA.
These characteristics within MTS-UDA render it a more complex and challenging task compared to standard TS UDA.

To address the above challenges, we propose SEnsor Alignment (SEA) for MTS-UDA. We first aim to reduce domain discrepancy at both the local and global sensor levels. At the local sensor level, we propose endo-feature alignment to align sensor-level information between domains. This alignment process includes aligning both the sensor features and sensor correlations, which capture the characteristics of individual sensors and their interactions, respectively. To achieve this alignment, we introduce contrastive sensor alignment and sensor correlation alignment. Further, at the global sensor level, we design exo-feature alignment to reduce global feature discrepancy. This is achieved by enforcing restrictions on the global features derived from stacked sensor features. Moreover, to capture the spatial-temporal dependencies inherent in MTS data and enable their effective transfer across domains, we introduce a graph-based encoder that is able to capture the dependency information. Given that MTS data may not come with pre-defined graphs that represent the relationships between sensors, we propose a multi-branch self-attention mechanism to model these dependencies.

{\color{black}Although SEA presents a feasible solution for sensor alignment in MTS-UDA, it employs a low-order statistics-based scheme for achieving endo-feature alignment. This scheme may be insufficient to achieve comprehensive alignment when dealing with complex data distributions in certain scenarios. Furthermore, SEA can benefit from leveraging the dynamically changing distributions within MTS data to enhance its performance. For instance, as a machine's health deteriorates, the distribution of a fan speed sensor undergoes dynamic changes due to increased friction. 
Taking these issues into consideration, we develop SEA++ as an extension of SEA by introducing multi-graph-based high-order alignment for enhancing our endo-feature alignment. It incorporates improved sensor correlation alignment and enhanced contrastive sensor alignment, utilizing high-order statistics to achieve comprehensive sensor-level alignment by capturing complex data distributions across domains. Additionally, SEA++ introduces a multi-graph alignment technique specifically tailored for aligning the evolving distributions within MTS data. This involves the construction of sequential graphs that represent the distributions within local temporal intervals, followed by the alignment of corresponding sequential graphs between domains. To optimize the alignment process, weights for balancing sequential graphs are learned based on distribution discrepancies, allowing us to prioritize alignment for challenging graphs. This module effectively aligns the evolving distributions, enabling adaptation to the changing nature of the data. With the two types of enhancements in the improved version, we represent the name with two plus signs as SEA++.}

Our contributions can be summarized as following.
\begin{itemize}
    \item We formulate a challenging scenario of Multivariate Time-Series Unsupervised Domain Adaptation (MTS-UDA) in accordance with the characteristics of MTS data. To our best knowledge, this is the first work to design a UDA method specifically for MTS data.
    \item We analyze the problems underlying MTS-UDA and propose SEnsor Alignment (SEA) that reduces domain discrepancy at both the local and global sensor levels. Meanwhile, SEA captures the spatial-temporal dependencies within MTS data, enabling effective transfer of the dependency information across domains.
    \item {\color{black}To cope with scenarios with complex data distributions, we introduce SEA++ by enhancing our endo-feature alignment. High-order statistics alignment methods are employed to achieve comprehensive sensor correlation and sensor feature alignment.}
    \item {\color{black}Considering the dynamically changing distributions within MTS data, SEA++ further incorporates a multi-graph alignment technique. It aligns sequential graphs representing local distributions, effectively aligning evolving distributions between domains.}
    \item We evaluate the effectiveness of SEA and SEA++ with various real-world MTS datasets through extensive experiments, demonstrating that the both methods achieve state-of-the-art performance for MTS-UDA.
\end{itemize}

{\color{black}This journal paper presents an extended version of our previous work \cite{wang2023sensor}  by introducing multi-graph-based high-order alignment techniques. First, we introduce high-order statistic alignment to improve endo-feature alignment, aiming to achieve more comprehensive sensor-level alignment. Second, we analyze the properties of MTS data and emphasize the significance of dynamically changing distributions inherent within MTS data. Thus, we introduce a multi-graph alignment technique to align the evolving distributions across domains. Third, we have conducted additional experiments, including more compared methods, comprehensive ablation studies, and extensive sensitivity analysis, to fully assess SEA and SEA++.}

\section{Related Work}
\label{related}
\subsection{Unsupervised domain adaptation}
UDA aims to mitigate the need for labeled data by transferring knowledge from a labeled source domain to an unlabeled target domain. To achieve good performance on the target domain, existing UDA methods have mainly focused on minimizing domain discrepancies by utilizing two categories of approaches.
The first category comprises adversarial-based methods, which utilize domain discriminator networks to encourage a feature extractor to learn domain-invariant representations. Prominent models in this category include Domain Adversarial Neural Network (DANN) \cite{10.5555/2946645.2946704} and Adversarial Discriminative Domain Adaptation (ADDA) \cite{8099799}. These models have served as foundations for various improved variants that address specific challenges \cite{long2018conditional,Pei_Cao_Long_Wang_2018,yu2019transfer,du2021cross,gao2021gradient,chen2022reusing}. For instance, CDAN \cite{long2018conditional} and MADA \cite{Pei_Cao_Long_Wang_2018} were both extensions of DANN that incorporate multi-mode structure information for classification tasks. DAAN \cite{yu2019transfer} introduced class-aware adversarial loss to align the conditional distribution using DANN. 
The second category encompasses metric-based methods, which enforce metric restrictions to enable the learning of invariant features. Commonly employed metrics include Maximum Mean Discrepancy (MMD) \cite{https://doi.org/10.48550/arxiv.1412.3474} and deep CORrelation ALignment (CORAL) \cite{10.1007/978-3-319-49409-8_35}. Within this category, several variants have been proposed to tackle specific challenges \cite{long2017deep,long2018transferable,chen2020homm,li2020domain}. For example, JAN \cite{long2017deep} was developed based on MMD to align the joint distribution of feature and label distributions. MK-MMD \cite{long2018transferable} employed MMD to align features in the final layers, enhancing their transferability. Furthermore, high-order moment matching called HoMM \cite{chen2020homm} was developed as a high-order moment matching technique to align feature distributions in hidden layers, addressing the limitations of low-order alignment methods. In addition, central moment discrepancy \cite{zellinger2017central} and maximum density divergence \cite{li2020maximum} are another two criterias to align feature distributions in hidden layers.
Overall, these methods have contributed to the advancement of UDA by effectively reducing the domain discrepancy and enabling knowledge transfer between different domains. 
\subsection{Unsupervised domain adaptation for time-series data}
In recent years, there has been a growing trend of exploring existing UDA methods in various domains, showing their effectiveness in different applications \cite{xu2023multi,xu2022aligning}. In the context of TS data, researchers have primarily focused on reducing domain discrepancy by aligning temporal features across different domains
\cite{Cai_Chen_Li_Chen_Zhang_Ye_Li_Yang_Zhang_2021,ijcai2021-378,Purushotham2017VariationalRA,https://doi.org/10.48550/arxiv.2203.08321,li2019domain}. 
Several notable works have made significant contributions to this area. For instance, VRADA \cite{Purushotham2017VariationalRA} leveraged a variational RNN to learn temporal features and employs adversarial-based methods to reduce the domain discrepancy. Similarly, Zhao et al. \cite{zhao2020deep} introduced a UDA method combining DANN and CNN for EEG classification, which incorporated a center loss to minimize intra-class variation and maximize inter-class distance. CoDATS \cite{wilson2020multi} and CLADA \cite{wilson2023calda}, as multi-source UDA methods for TS data, applied DANN and 1D-CNN to handle the scenarios where multiple source domains are available.
Moreover, some researchers have specifically designed UDA methods tailored to TS data. ADATIME \cite{https://doi.org/10.48550/arxiv.2203.08321} considered the temporal properties of TS data and evaluated various temporal encoders for TS UDA. Besides, AdvSKM \cite{liu2021adversarial} mapped features into a high-dimensional space by proposing adversarial Spectral Kernels and then aligned them using MMD. To consider the invariant associative structure between variables exhibited by MTS data, SASA \cite{Cai_Chen_Li_Chen_Zhang_Ye_Li_Yang_Zhang_2021} aligned this structural information for TS UDA.

These works are great pioneers for TS UDA, but they may not perform well for the scenarios requiring multiple sensors, i.e., MTS data. In MTS data, signals from different sensors often follow various distributions due to their varied physical locations and measurements. Existing TS UDA methods can be adapted to MTS data by treating all sensors as a whole \cite{YANG2021104498,https://doi.org/10.48550/arxiv.2203.08321,Purushotham2017VariationalRA,zhao2020deep}, facilitate the learning of global features for alignment, but they are limited in accounting for the distinct distributions of sensors within MTS data. Meanwhile, they ignored the multi-source nature of MTS data, restricting their ability to model and transfer spatial-temporal dependency information across domains. 
To address these limitations, we propose our SEA and SEA++ that accounts for aligning sensor-level information and modelling dependency information within MTS data to achieve better MTS-UDA.




\begin{figure*}[!ht]
    \centering
    \includegraphics[width = 1\linewidth]{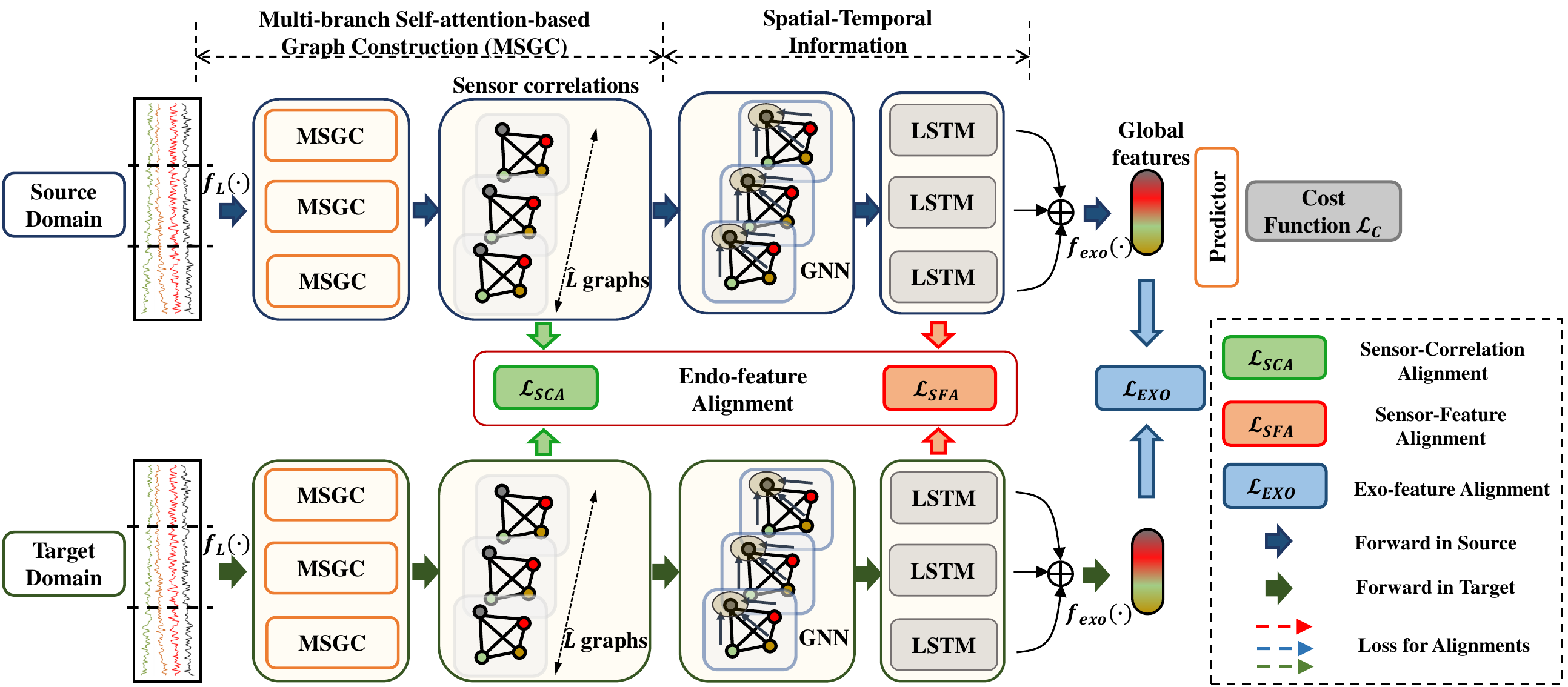}
    \caption{The overall structure. (1) To transfer spatial-temporal dependencies across domains, source and target domains share the same graph-based encoder, including MSGC, GNN, and LSTM. Each sample is segmented as multiple patches, which are constructed as sequential graphs. With the combination of GNN and LSTM, decent sensor information is learned, including sensor features and correlations. (2) In addition to supervised learning on the source domain, endo-feature alignment and exo-feature alignment are designed to reduce domain discrepancy at the local and global sensor levels. The endo-feature alignment includes aligning sensor features and their correlations between domains by SCA and SFA, both of which are enhanced by incorporating multi-graph alignment and high-order statistic alignment. Exo-feature alignment aligns global features mapped by sensor features.}
    \label{fig:overall}
\end{figure*}

\section{SEA Model}
\label{method}
\subsection{Problem Definition}
For MTS-UDA, we are given a source domain with $N_s$ labeled samples $\mathcal{D}_S = \{(x_i^s, y_i^s)\}_{i=1}^{N_s}$ and a target domain with $N_t$ unlabeled samples $\mathcal{D}_T = \{x_i^t\}_{i=1}^{N_t}$. Each MTS sample $x_i$ (either $x_i^s$ or $x_i^t$) originates from $N$ sensors with different distributions, i.e., $x_i = \{x_{im}\}_{m=1}^N\in\mathbb{R}^{N\times{L}}$, where $L$ represents the time length. The goal of SEA is to train an encoder by transferring the knowledge from the source domain to the target domain, enabling the effective learning of features $h_i\in\mathbb{R}^{F}$ from the MTS data $x_i^t$ in the target domain. The features can then be used for downstream tasks such as RUL prediction and HAR. 
Notably, the SEA framework consists of two stages. In the first stage, we focus on feature extraction from MTS data. To simplify the notation and enhance clarity, we omit the index $i$ in this stage. Thus, each sample is represented as $x^s$ and $x^t$, and the data from the $m$-th sensor is denoted as $x_m^s$ and $x_m^t$.
While in the second stage that involves domain alignment, we retain the use of the index to facilitate a clear description of the alignment process.

Furthermore, to model the evolving distributions within MTS data, we construct sequential graphs $\{\mathcal{G}_T\}_{T=1}^{\hat{L}}, \mathcal{G}_T = (Z_T, E_T)$ from the sample $x$, where $Z_T=\{z_{m,T}\}_{m=1}^N$ and $E_T=\{e_{mn, T}\}_{m, n=1}^N$ represent the sensor features and correlations, respectively, in the $T$-th graph. 
$\{\mathcal{G}^{s}_T\}_{T=1}^{\hat{L}}$ and $\{\mathcal{G}^{t}_T\}_{T=1}^{\hat{L}}$ represent the sequential graphs in the source and target domains respectively. 


\subsection{Preliminary}
\subsubsection{Graph Neural Network}

Several previous studies have demonstrated the effectiveness of Graph Neural Network (GNN) in capturing spatial dependencies \cite{deng2021graph, ijcai2020-184}. Motivated by this capability, we employ GNN to capture the spatial dependencies, i.e., sensor correlations, within our constructed sequential graphs. Specifically, we utilize a variant of GNN known as Message Passing Neural Network (MPNN) \cite{gilmer2017neural} to process each graph. The MPNN framework consists of two stages: propagation and updating. During the propagation stage, the features of neighboring nodes are propagated to the central node. Subsequently, the aggregated features are updated using a non-linear function, such as a Multi-Layer Perceptron (MLP) network. With MPNN, we can capture the spatial dependencies between sensors, enabling the effective transfer of the dependency information across domains in the subsequent alignment processes. The details of MPNN are shown in Eq. (\ref{eq:mpnn}), where sensor $m$ is the central node, and $\mathcal{N}(m)$ is the set of its neighboring nodes.
\begin{equation}
    \label{eq:mpnn}
    \begin{split}
        {h_{m, T}} &= \sum_{j\in\mathcal{N}(m)}{e_{mj, T}}{z_{j, T}}, \\
        {z_{m, T}} &= ReLU({h_{m, T}}W_{G}).
    \end{split}
\end{equation}
\subsubsection{Deep Coral}
Deep Coral \cite{10.1007/978-3-319-49409-8_35} is one of the most typical metric-based methods for UDA. Different from $\mathcal{L}_2$ or cosine distances which mainly employ first-order statistics, Deep Coral focuses on minimizing the discrepancy in second-order statistics between domains \cite{guo2020multi}, making it possible to capture complex data distributions for alignment. Given training batches with $n^s$ and $n^t$ samples in source and target domains respectively, we denote the extracted features from these domains as $\mathcal{H}^s\in\mathbb{R}^{n^s\times{f}}$ and $\mathcal{H}^t\in\mathbb{R}^{n^t\times{f}}$, where $f$ represents the feature dimension. With the features, we can compute the discrepancy of second-order statistics with Deep Coral $M_c(\mathcal{H}^s,\mathcal{H}^t)$ as Eq. (\ref{eq:coral}), where $||\cdot||_F^2$ is the squared matrix Frobenius norm. By minimizing the Deep Coral discrepancy $\mathcal{L}_{Coral}$, the second-order statistics of the features are aligned between the source and target domains, leading to comprehensive alignment across domains.
\begin{equation}
    \label{eq:coral}
    \begin{split}
        &C^s = \frac{(\mathcal{H}^s)^T\mathcal{H}^s - \frac{1}{n^s}(\textbf{1}^T\mathcal{H}^s)(\textbf{1}^T\mathcal{H}^s)}{n^s-1},\\
        &C^t = \frac{(\mathcal{H}^t)^T\mathcal{H}^t - \frac{1}{n^t}(\textbf{1}^T\mathcal{H}^t)(\textbf{1}^T\mathcal{H}^t)}{n^t-1},\\
        & M_c(\mathcal{H}^s,\mathcal{H}^t) = \frac{1}{4f^2}||C^s-C^t||_F^2.
    \end{split}
\end{equation}

\subsection{Overall Structure}
\label{sec:Overall}
The overall structure of SEA is shown in Fig. \ref{fig:overall}. The superscripts $s$ and $t$ are omitted in Section \ref{sec:Overall} and \ref{sec:MSGC}, as the forward processes in source and target domains share the same graph-based encoder, including Multi-branch Self-attention based Graph Construction (MSGC), GNN, and LSTM. First, we model the spatial-temporal dependencies within MTS data for effective transfer across domains. Given the sample $x$, we segment it into multiple patches, which are constructed as sequential graphs by MSGC. Then, with the sequential graphs, GNN and LSTM are leveraged to capture the spatial-temporal information for learning sensor-level information, i.e., sensor features and correlations. Second, in addition to supervised learning on the source domain, we incorporate endo-feature alignment and exo-feature alignment techniques to reduce domain discrepancy at both the local and global sensor levels. At the local sensor level, we design endo-feature alignment, which involves aligning sensor features and sensor correlations by proposed Sensor Feature Alignment (SFA) and Sensor Correlation Alignment (SCA) respectively. {\color{black}The two components are further improved by both the high-order statistic alignment and multi-graph alignment to form iSFA and iSCA. The high-order statistic alignment aims to achieve comprehensive alignment by capturing complex data distributions while the multi-graph alignment aims to align evolving distributions at the sensor level across domains.}
At the global sensor level, we design exo-feature alignment to align global features by enforcing restrictions. 
We introduce the details of each module in subsequent sections.

\subsection{Graph-based Encoder}
\label{sec:MSGC}
\subsubsection{Segmentation}

To capture the evolving distributions within MTS data, we employ a segmentation approach. In this approach, a sample $x\in\mathbb{R}^{N\times{L}}$ is segmented into multiple patches, each with a size of $d$, yielding sequential features denoted as $\{Z_T\}_{T=1}^{\hat{L}}\in\mathbb{R}^{N\times{\hat{L}\times{d}}}$. Here, $\hat{L} = \lfloor{\frac{L}{d}}\rfloor$ represents the number of patches, and $d$ represents the feature dimension. The features within the $T$-th patch $Z_T\in\mathbb{R}^{N\times{d}}$ capture the distributions in local temporal intervals. Within each patch, the features of the $m$-th sensor are denoted as $z_{m, T}\in\mathbb{R}^d$. Additionally, we incorporate a nonlinear function $f_L(\cdot)$ to enhance the nonlinear expressiveness of the model.


\subsubsection{MSGC}
We construct sequential graphs with the sequential features. 
{\color{black}Traditionally, various methods have been employed to construct graphs for MTS data, such as dot-product\cite{wang2023multivariate}, Euclidean distance\cite{ijcai2020-184}, and concatenation \cite{cirstea2019graph}. However, the traditional methods are insufficient to model comprehensive relationships of sensors due to the complex topological structures presented within MTS data.}
To solve the limitation, we propose MSGC, {\color{black}which constructs a graph by combining multiple graphs generated from various weight initializations across multiple branches. Each branch employs differently weighted matrices to model sensor correlations from distinct perspectives, allowing us to capture comprehensive sensor correlations.} Specifically, we adopt query and key to learn the edge weights between sensors. Eq. (\ref{eq:multibranch}) describes the edge between sensors $m$ and $n$ in the $i$-th branch.
\begin{equation}
    \label{eq:multibranch}
    \begin{split}
        q_{m,T}^i &= z_{m, T}W_Q^i, k_{n, T}^i = z_{n, T}W_K^i,\\
        e_{mn, T}^i &= \frac{\frac{q_{m, T}^i(k_{n, T}^i)^T}{\sqrt{d}}}{\sum_{j=1}^N\frac{q_{m, T}^i(k_{j, T}^i)^T}{\sqrt{d}}}, \\
    \end{split}
\end{equation}
where we treat the $m$-th sensor $z_{m, T}$ as the query sensor and the $n$-th sensor $z_{n, T}$ as the key sensor, and use this information to learn the edge weight between them for the $i$-th branch, i.e., $e_{mn, T}^i$. The edge weights are then normalized using the softmax function to ensure they fall within the range of [0, 1].

To combine the graphs from multiple branches, we calculate the average of the corresponding elements, i.e., $e_{mn, T} = \sum_{i=1}^{n_{b}}\frac{e_{mn, T}^i}{n_{b}}$, where $n_b$ represents the number of branches. This operation allows us to obtain the sensor correlations $E_T = \{e_{mn, T}\}_{m,n=1}^N\in\mathbb{R}^{N\times{N}}$ for the $T$-th graph. By conducting similar operations in other patches, we can finally obtain the sequential correlations $\{E_T\}_{T=1}^{\hat{L}}\in\mathbb{R}^{\hat{L}\times{N}\times{N}}$.



\begin{figure*}[t]
    \centering
    \includegraphics[width = .9\linewidth]{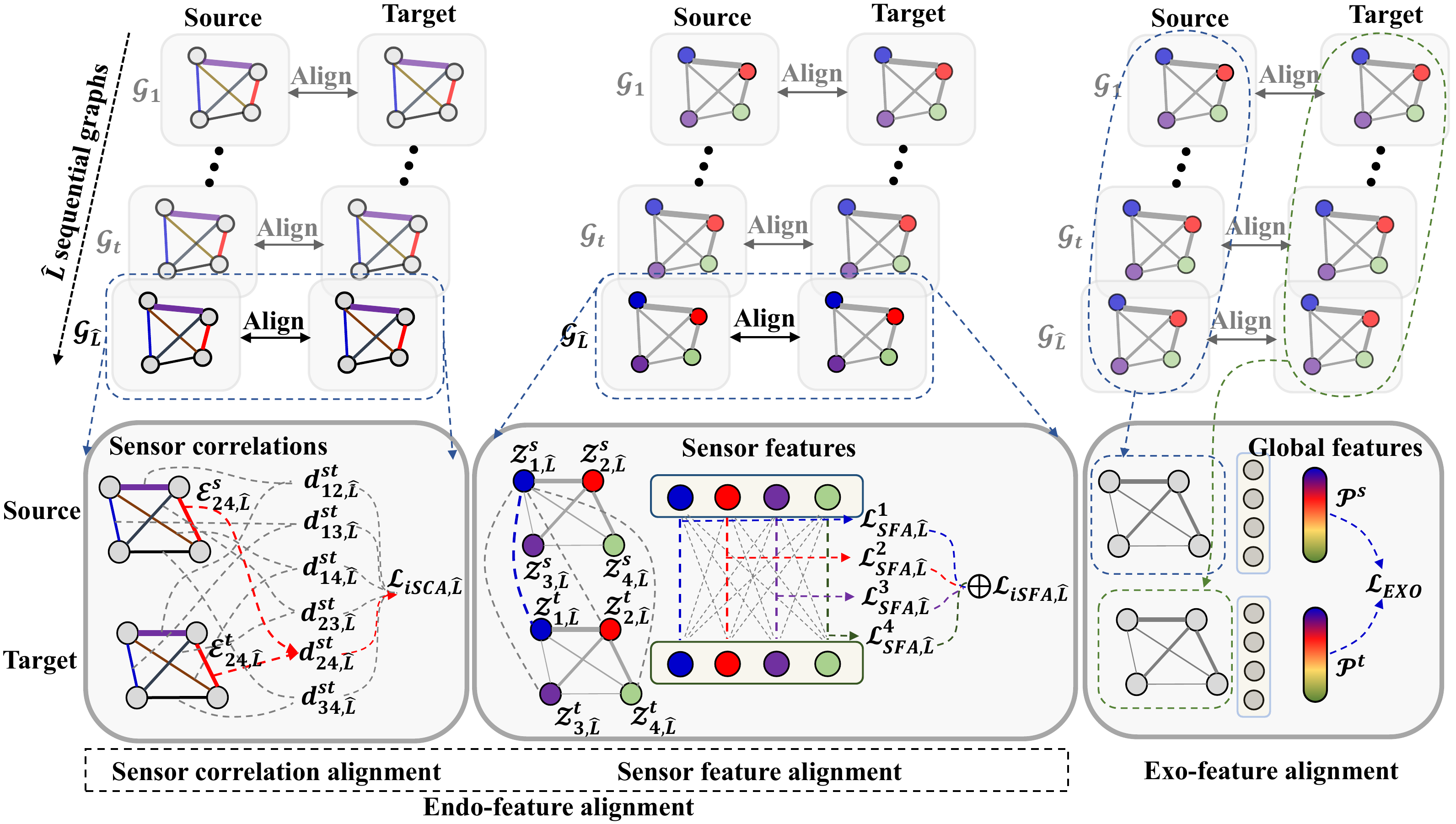}
    \caption{{Left}: Sensor correlation alignment is to make the correlations between domains identical. {Middle}: Sensor feature alignment is to make the features from the corresponding sensors in two domains similar and the data from the different sensors in two domains different. {Right}: Exo-feature alignment reduces the discrepancy between the global features in different domains.}
    \label{fig:corralign}
\end{figure*}

\subsubsection{Capturing Spatial-Temporal Information}
\label{sec:ST}
With the constructed sequential graphs, we employ two models, MPNN and LSTM, to capture the spatial-temporal dependencies within MTS data. First, we utilize MPNN (Eq. (\ref{eq:mpnn})) to capture the sensor correlations within each sequential graph. In this process, the graphs are processed with shared weights to save training costs and trainable parameters. Once the features of each sensor in the sequential graphs have been updated using MPNN, our focus shifts to capture temporal dependencies. It is noted that features from corresponding sensors across the sequential graphs show temporal dependencies. To capture these dependencies, we employ LSTM, which captures the dependency information for each sensor by taking its features $z_m\in\mathbb{R}^{{\hat{L}}\times{d}}$ across sequential graphs.
After GNN and LSTM, we finally obtain updated sequential features $\{Z_T\}_{T=1}^{\hat{L}}\in\mathbb{R}^{\hat{L}\times{N}\times{d}}$. By combining the capabilities of the two networks, we can effectively capture the spatial-temporal dependencies within MTS data, allowing us to transfer the dependency information across domains through the alignment at the local and global sensor levels in the subsequent stages.

\subsection{SEnsor Alignment}

To reduce the discrepancy between domains for MTS-UDA, we design endo-feature alignment and exo-feature alignment at the local and global sensor levels respectively. 
\subsubsection{Endo-feature alignment}
Existing UDA methods reduce domain discrepancy by aligning global features across domains, which is applicable for the data coming from a single source. However, MTS data originate from multiple sensors, with each sensor following various distributions. When adapting existing UDA methods to MTS-UDA, they can consider global distributions by treating all sensors as a whole yet ignore the distinct distributions of each sensor. This results in the misalignment at the sensor level, restricting the performance of the learned model transferring from the source domain to the target domain. To address this limitation, we propose an endo-feature alignment method designed to align sensor-level information and avoid misalignment for each sensor when transferring knowledge across domains.

To make better endo-feature alignment, it is necessary to note that the sensor-level information consists of both the sensor features and sensor correlations. Sensor features represent the properties of each sensor, while sensor correlations represent the interactive information between sensors. For illustration, we present a scenario with temperature and fan speed sensors deployed together to monitor the status of a machine. 
It is desirable for the data from temperature sensors in two domains to follow similar distributions. Additionally, the correlation between the sensors should also demonstrate consistent patterns across domains, such as the fan speed consistently increasing with temperature rises. Based on these considerations, aligning sensor features and sensor correlations across domains simultaneously is important to ensure comprehensive alignment at the sensor level.

\paragraph{Sensor Correlation Alignment}
To explore consistent patterns within the interactions of sensors across domains, we propose to align the sensor correlations. Specifically, our aim is to align the distribution of correlations between any two sensors across domains. To achieve this, we propose sensor correlation alignment (SCA), which involves individually aligning each edge of the graphs as shown in Fig. \ref{fig:corralign} \textbf{Left}.

{\color{black}Given training batches with $n^s$ and $n^t$ samples in source and target domains, we represent sensor correlations across sequnetial graphs as $\mathcal{E}^s\in\mathbb{R}^{n^s\times{\hat{L}}\times{N}\times{N}}$ and $\mathcal{E}^t\in\mathbb{R}^{n^t\times{\hat{L}}\times{N}\times{N}}$ respectively, with $\mathcal{E}_T^s\in\mathbb{R}^{n^s\times{N}\times{N}}$ and $\mathcal{E}_T^t\in\mathbb{R}^{n^t\times{N}\times{N}}$ signifying sensor correlations for the $T$-th graph. Here, $\mathcal{E}_T^s = \{\mathcal{E}_{mn,T}^s\}_{m,n=1}^N$, wherein $\mathcal{E}_{mn,T}^s = \{e^s_{i,mn,T}\}_{i}^{n^s}$. For SCA, alignment of sensor correlations across domains is achieved based on averaged sensor correlations. Specifically, for the edge connecting sensors $m$ and $n$, its mean value across the training batch is computed as $e^s_{mn,T} = \sum_i^{n^s}e^s_{i,mn,T}/n^s$. Subsequently, the averaged sensor correlations over sequential graphs are calculated as $e^s_{mn} = \sum_T^{\hat{L}}e^s_{mn,T}/{\hat{L}}$.} Next, we minimize the averaged sensor correlations across domains as shown in Eq. (\ref{eq:SCA}), where $|\cdot|$ is absolute value. This involves minimizing the expected discrepancy of sensor correlations between the source and target domains. By doing so, we can reduce distribution discrepancies for sensor correlations in source and target domains. 
\begin{equation}
    \label{eq:SCA}
    \begin{split}
        e_{mn}^{st} &= e_{mn}^s-e_{mn}^t,\\
        \mathcal{L}_{SCA} &= \mathbb{E}(|e_{mn}^{st}|).
    \end{split}
\end{equation}
\paragraph{Sensor Feature Alignment}
Next, we focus on aligning sensor features between domains. Our assumption is that features originating from corresponding sensors in two domains should exhibit similar distributions. Meanwhile, compared with features from corresponding sensors, features from different sensors in two domains should have distinct properties. 
For example, as shown in Fig. \ref{fig:corralign} \textbf{Middle}, the distribution of sensor 1 in the source domain should be more similar to the distribution of its corresponding sensor in the target domain than to the distributions of other sensors (e.g., sensors 2, 3, and 4) in the target domain.
To achieve this goal, we propose a sensor contrasting mechanism that facilitates the alignment of sensor features between domains.

{\color{black}Given training batches with $n^s$ and $n^t$ samples, we define sensor features as $\mathcal{Z}^s\in\mathbb{R}^{n^s\times{\hat{L}}\times{N}\times{d}}$ and $\mathcal{Z}^t\in\mathbb{R}^{n^t\times{\hat{L}}\times{N}\times{d}}$, with $\mathcal{Z}_T^s\in\mathbb{R}^{n^s\times{N}\times{d}}$ and $\mathcal{Z}_T^t\in\mathbb{R}^{n^t\times{N}\times{d}}$ denoting sensor features for the $T$-th graph. Here, $\mathcal{Z}_T^s = \{\mathcal{Z}_{n,T}^s\}_{n=1}^N$, wherein $\mathcal{Z}_{n,T}^s = \{z^s_{i,n,T}\}_{i}^{n^s}$. In SFA, domain contrasting is realized based on averaged sensor features. In Eq. (\ref{eq:SFA}), $p^s_m$ represents the averaged sensor features over multiple graphs within a training batch, calulated as $p^s_m = \sum_T^{\hat{L}}\sum_i^{n^s}z^s_{i,n,T}/n^s\hat{L}$.}
\begin{equation}
    \label{eq:SFA}
    \begin{split}
        \mathcal{L}_{SFA} &= -\frac{1}{N}\sum_n^Nlog\frac{e^{\varphi(p_m^s, p_m^t)}}{\sum_{p_j^t\in{P^t}}e^{\varphi(p_m^s, p_j^t)}},\\
        \varphi(p_m^s, p_m^t) &= p_m^s(p_m^{t})^T.
    \end{split}
\end{equation}
\paragraph{Exo-feature alignment.}
To reduce the domain discrepancy at the global level, we enforce restrictions on the global features mapped by sensor features between domains. Specifically, we first need to learn the global features of each sample. This involves stacking all sensor features across the sequential graphs and subsequently employing an MLP network to map the stacked features as global features $\mathcal{P}^s$ and $\mathcal{P}^t$ in source and target domains, as shown in Fig. \ref{fig:corralign} \textbf{Right}. With the global features, we can then align them with Deep Coral, expressed as $\mathcal{L}=\mathcal{M}_c(\mathcal{P}^s,\mathcal{P}^t)$.
Finally, the overall loss function (\ref{eq:overloss}) is minimized:
\begin{equation}
    \label{eq:overloss}
    \begin{split}
        \min\mathcal{L} = \mathcal{L}_C + \mathcal{L}_{EXO} + \mathcal{L}_{Endo},\\
        \mathcal{L}_{Endo} =  \lambda_{SCA}\mathcal{L}_{SCA} + \lambda_{SFA}\mathcal{L}_{SFA}.
    \end{split}
\end{equation}
where $\mathcal{L}_C$ is the cost function (such as mean square error loss or cross-entropy loss, depending on specific tasks) computed by the source data and labels, and $\lambda_{SCA}$ and $\lambda_{SFA}$ are hyperparameters to balance the effect of SCA and SFA respectively.

{\color{black}\section{SEA++ Model}
While the method described above offers a viable approach for achieving sensor correlation and sensor feature alignment, there is still room for improvement to enhance endo-feature alignment. Two key directions warrant consideration.
First, SEA employs first-order statistics for alignment, such as averaged sensor features over a training batch used in the sensor contrastive mechanism. However, the first-order statistical alignment method is limited in dealing with scenarios with complex data distributions. This limitation may restrict the ability to achieve comprehensive alignment across domains, leading to suboptimal performance in certain scenarios. To address that, we propose introducing second-order statistical metrics for alignment to achieve more comprehensive endo-feature alignment.

Second, the constructed sequential graphs, comprising sequential features and correlations, capture evolving distributions within local temporal intervals and should be aligned across domains.
We consider the example of monitoring machine health status for illustration. As the machine's health deteriorates, the distribution of a fan speed sensor may change due to increased friction or other factors. Similarly, the interaction between the fan speed sensor and a temperature sensor will also be affected and undergo changes. Specifically, the same fan speeds might lead to higher temperature increases due to a machine in poor health, e.g., friction increasing. Thus, we emphasize the importance of aligning the evolving distributions at the sensor level between domains. However, SEA employs averaged sensor correlations and sensor features over the sequential graphs, limiting its ability to capture the evolving distributions for alignment. To improve this, we introduce a multi-graph alignment technique, aiming to effectively align evolving distributions between domains.

Next, we begin by discussing SCA and SFA enhanced by high-order statistical alignment within each sequential graph. Subsequently, we focus on combining the alignment across all sequential graphs by employing our multi-graph alignment.

\subsection{Sensor Correlation Alignment}
Given training batches with $n^s$ and $n^t$ samples in source and target domains, we have sensor correlations for the $T$-th graph, denoted as $\mathcal{E}_T^s\in\mathbb{R}^{n^s\times{N}\times{N}}$ and $\mathcal{E}_T^t\in\mathbb{R}^{n^t\times{N}\times{N}}$ respectively. Instead of leveraging averaged sensor correlations within batches, we introduce second-order statistics with Deep Coral, i.e., $M_{c}$ in Eq. (\ref{eq:coral}), to align the sensor correlations between the source and target domains. Building upon this idea, we enhance SCA and introduce improved SCA (iSCA) as shown in Eq. (\ref{eq:iSCA}).
\begin{equation}
    \label{eq:iSCA}
    \begin{split}
        d_{mn,T}^{st} &= M_c(\mathcal{E}_{mn,T}^s,\mathcal{E}_{mn,T}^t),\\
        \mathcal{L}_{iSCA,T} &= \mathbb{E}(d_{mn,T}^{st}) = \frac{\sum_{m}^N\sum_n^Nd_{mn,T}^{st}}{mn}.
    \end{split}
\end{equation}
By aligning each edge between domains with the second-order statistical metric and minimizing the distribution discrepancy of sensor correlations, we can achieve more comprehensive alignment for sensor correlations.


\subsection{Sensor Feature Alignment}
Next, we focus on improving sensor feature alignment. Similar to SCA, the sensor contrasting mechanism in SFA is realized with the averaged sensor features over the training batches in source and target domains. Thus, SFA encounters a challenge similar to SCA. Specifically, the contrasting-based alignment procedure in Eq. (\ref{eq:SFA}) relies on first-order statistics, which might not achieve comprehensive alignment when dealing with complex distributions. To overcome this limitation, we enhance SFA and introduce improved SFA (iSFA) by incorporating Deep Coral into the contrasting process.

In Eq. (\ref{eq:iSFA}), iSFA begins by calculating sensor-level distribution discrepancies using Deep Coral, which takes into account second-order statistics. Subsequently, iSFA leverages the sensor-level discrepancies to perform contrasting between domains. This involves minimizing the distribution discrepancies between corresponding sensors across domains while maximizing the distribution discrepancies between different sensors across domains. This approach allows for more comprehensive alignment of sensor features.
\begin{equation}
    \label{eq:iSFA}
    \begin{split}
        \mathcal{L}_{iSFA,T} = -\frac{1}{N}\sum_n^Nlog\frac{e^{M_c(\mathcal{Z}_{n,T}^s,\mathcal{Z}_{n,T}^t})}{\sum^N_{j}e^{M_c(\mathcal{Z}_{n,T}^s,\mathcal{Z}_{j,T}^t)}}.
    \end{split}
\end{equation}
\subsection{Multi-Graph Alignment}
To combine the alignment for sensor correlations and sensor features across all sequential graph, a straightforward approach would involve computing the average alignment for these graphs. However, the distributions between domains for different graphs might exhibit various discrepancies. To account for this and enhance domain adaptation, we propose a method to learn adaptive weights for adjustment, as shown in Eq. (\ref{eq:multi_align}).
\begin{equation}
    \label{eq:multi_align}
    \begin{split}
        \mathcal{L}_{iEndo} &= \sum_T\mathcal{W}_T(\lambda_{SCA}\mathcal{L}_{iSCA,T}+\lambda_{SFA}\mathcal{L}_{iSFA,T}),\\
        \mathcal{W}_T &= M_c([\mathcal{Z}_{1,T}^s,...,\mathcal{Z}_{\hat{L},T}^s],[\mathcal{Z}_{1,T}^t,...,\mathcal{Z}_{\hat{L},T}^t]),
    \end{split}
\end{equation}
where $[a,...,b]$ represents concatenating the elements from $a$ to $b$. In Eq. (\ref{eq:multi_align}), the learnable weights $\mathcal{W}_T$ are determined by evaluating the distribution discrepancy of the $T$-th graph between domains. Specifically, $\mathcal{W}_T$ will be large when there is a substantial discrepancy in the distributions of the $T$-th graph between the source and target domains. By assigning larger weights to the graphs with larger distribution discrepancies between domains, we prioritize alignment in those graphs. This adaptive weighting scheme allows us to better adapt the model to the evolving distributions between domains. Furthermore, we also introduce $\lambda_{SCA}$ and $\lambda_{SFA}$ to balance the effect of iSCA and iSFA.
By combining the exo-feature alignment and the improved endo-feature alignment, the improved overall loss function (\ref{eq:overlossi}) can be obtained as:
\begin{equation}
    \label{eq:overlossi}
    \begin{split}
        \min\mathcal{L} &= \mathcal{L}_C + \mathcal{L}_{EXO} + \mathcal{L}_{iEndo}.
    \end{split}
\end{equation}}

\section{Experiment}
\label{exp}
\subsection{Datasets and Setup}
To evaluate the performance of SEA and SEA++ under various MTS scenarios, we conducted experiments on two public datasets: C-MAPSS for remaining useful life prediction \cite{4711414}, which collects MTS data from different sensor types, and Opportunity HAR for human activity recognition \cite{5573462}, which collects MTS data from different locations.

\textbf{C-MAPSS} describes the degradation process of aircraft engines. The MTS data in C-MAPSS were originated from 14 sensors to measure various physical parameters, such as temperature, fan speed, and pressure. The dataset includes four sub-datasets collected under different working conditions and fault modes, where each sub-dataset represents one domain.
We processed the sub-datasets following the data preparation in the previous work \cite{9234721}, and data annotations represent the remaining useful life circle of engines.


\textbf{Opportunity HAR} describes human activities. The MTS data in Opportunity HAR were originated from 113 sensors deployed to various locations of the human body, such as hand and leg. After removing the sensors with constant values, 110 sensors were used for evaluation. The dataset includes the data collected from four subjects, each representing one domain. The data annotations include two levels: 1). locomotion representing the low-level tasks including four categories, sitting, standing, walking, and lying down; 2). gestures representing the high-level tasks including 17 various actions. Following the experimental settings in the previous work \cite{9804766}, we adopted the low-level tasks. As some values in the data were missing, we adopted the linear interpolation approach to fill in the missing positions. To construct the training dataset, we adopted a sliding window with a size of 128 and an overlapping of 50\% following the previous work \cite{9804766}.

The experiments include three parts, comparisons with state-of-the-art methods, the ablation study, and sensitivity analysis. All experiments were run ten times with average results reported to remove the effect of random initialization. Besides, we set the batch size as 50 and employed the optimizer as Adam with a learning rate of 0.001. Meanwhile, we adopted 20 training epochs for training our model. Furthermore, we built and trained our model based on Pytorch 1.9 and NVIDIA GeForce RTX 3080Ti GPU. 

We employed different evaluation indicators to assess the performance of our method on the two distinct datasets. For C-MAPSS experiments, which involve predicting the RUL of an engine, we utilized Root Mean Square Error (RMSE) and the Score function \cite{9234721, 8998569, 9351733,fu2021novel} as evaluation metrics. Lower values for both indicators indicate better model performance. For Opportunity HAR experiments, which entail a classification task, we employed accuracy as the evaluation metric. High accuracy values signify superior model performance.

\subsection{Comparisons with State-of-the-Art}
\begin{table*}[htbp]
  \centering
      \caption{The Comparisons with SOTAs in C-MAPSS (\textbf{R}: RMSE; \textbf{S}: Score)}
    \begin{tabular}{lccccccccccccc}
    \toprule
    \toprule
    Models (\textbf{R}) & \textbf{1$\to$2} & \textbf{1$\to$3} & \textbf{1$\to$4} & \textbf{2$\to$1} & \textbf{2$\to$3} & \textbf{2$\to$4} & \textbf{3$\to$1} & \textbf{3$\to$2} & \textbf{3$\to$4} & \textbf{4$\to$1} & \textbf{4$\to$2} & \textbf{4$\to$3} & \textbf{Avg.} \\
    \midrule
    {\color{black}Target} & {\color{black}12.93} & {\color{black}13.33} & {\color{black}14.66} & {\color{black}12.36} & {\color{black}13.33} & {\color{black}14.66} & {\color{black}12.36} & {\color{black}12.93} & {\color{black}14.66} & {\color{black}12.36} & {\color{black}12.93} & {\color{black}13.33} & {\color{black}13.32} \\
    Source & 28.76 & 35.28 & 27.21 & 21.86 & 39.80 & 33.50 & 37.90 & 30.33 & 24.10 & 33.76 & 25.31 & 21.75 & 29.96 \\
    \midrule
    DDC   & 43.25 & 39.48 & 42.99 & 40.07 & 39.46 & 43.01 & 40.93 & 43.43 & 43.72 & 41.58 & 43.38 & 39.61 & 41.74 \\
    Coral & 16.72 & 26.49 & 25.03 & 14.47 & 37.41 & 31.13 & 32.82 & 23.40 & 19.84 & 28.37 & 32.00 & 22.47 & 25.85 \\
    DANN  & 17.02 & 23.33 & 24.20 & 14.16 & 30.00 & {25.91} & {19.89} & 20.31 & 19.95 & 31.74 & 24.68 & 17.94 & 22.43    \\
    BNM   & 48.18 & 41.61 & 46.45 & 56.66 & 48.40 & 62.02 & 49.39 & 48.89 & 42.98 & 40.76 & 43.71 & 41.89 & 47.58 \\
    SDAT  & 16.50 & 24.11 & \underline{23.07} & 13.57 & 29.76 & 26.38 & 25.34 & 20.29 & \underline{18.63} & 20.13 & 23.23 & 17.65 & 21.55 \\
    {\color{black}AdvSKM} & {\color{black}31.09} & {\color{black}34.16} & {\color{black}40.74} & {\color{black}23.93} & {\color{black}32.41} & {\color{black}35.73} & {\color{black}28.47} & {\color{black}22.17} & {\color{black}26.45} & {\color{black}28.56} & {\color{black}22.75} & {\color{black}23.51} & {\color{black}29.16} \\
    {\color{black}CoDATs} & {\color{black}15.91} & {\color{black}26.67} & {\color{black}25.24} & {\color{black}15.59} & {\color{black}\textbf{24.52}} & {\color{black}25.64} & {\color{black}21.25} & {\color{black}21.04} & {\color{black}19.77 }& {\color{black}19.41} & {\color{black}18.99} & {\color{black}17.85} & {\color{black}20.99} \\
    CtsADA & 18.31 & 29.23 & 29.49 & 23.20 & 54.69 & 41.60 & 38.14 & 24.67 & 19.86 & 31.63 & 25.71 & 23.97 & 30.04 \\
    CLUDA & 25.31 & 36.80 & 33.24 & 22.51 & 33.82 & 34.63 & 26.48 & 24.09 & 26.29 & 24.33 & 25.14 & 22.18 & 27.90 \\
        \midrule
    SEA   & \underline{15.63} & \underline{21.09} & 23.21 & \underline{13.29} & {24.62} & \underline{25.32} & \underline{19.69} & \underline{19.44} & 19.37 & \underline{18.92} & \textbf{17.70} & \textbf{16.45} & \underline{19.56} \\
    SEA++  & \textbf{15.41} & \textbf{19.38} & \textbf{22.12} & \textbf{13.00} & \underline{24.60} & \textbf{24.90} & \textbf{19.19} & \textbf{19.25} & \textbf{17.75} & \textbf{18.43} & \underline{18.16} & \underline{16.79} & \textbf{19.08} \\
    \midrule
    \midrule
    Models (\textbf{S}) & \textbf{1$\to$2} & \textbf{1$\to$3} & \textbf{1$\to$4} & \textbf{2$\to$1} & \textbf{2$\to$3} & \textbf{2$\to$4} & \textbf{3$\to$1} & \textbf{3$\to$2} & \textbf{3$\to$4} & \textbf{4$\to$1} & \textbf{4$\to$2} & \textbf{4$\to$3} & \textbf{Avg.} \\
    \midrule
    {\color{black}Target} & {\color{black}678}   & {\color{black}330}   & {\color{black}1027}  & {\color{black}241}   & {\color{black}330}   & {\color{black}1027}  & {\color{black}241}   & {\color{black}678}   & {\color{black}1027}  & {\color{black}241}   & {\color{black}678}   & {\color{black}330}   & {\color{black}569} \\
    Source & 9341  & 4751  & 4699  & 2340  & 8335  & 11018 & 24892 & 26878 & 11006 & 16318 & 8771  & 2047  & 10866 \\
    \midrule
    DDC   & 23323 & 6164  & 34891 & 11581 & 6578  & 35122 & 22236 & 59400 & 80774 & 29592 & 58029 & 15239 & 31911 \\
    Coral & 1158  & 1468  & 3613  & 406   & 6241  & 8492  & 15298 & 12661 & 3048  & 6742  & 25763 & 4758  & 7471 \\
    DANN  & 1073  & 1612  & 3340  & 392   & 3160  & 4667  & 1666   & 3896  & 2381  & 22225 & 4620  & 762   & 4150 \\
    BNM   & 24265 & 44650 & 19017 & 526203 & 56769 & 29689 & 46262 & 29188 & 39559 & 11618 & 20566 & 65807 & 76133 \\
    SDAT  & 974   & 1403  & \underline{2729}  & 300   & 4360  & 3859  & 3756  & 3134  & 2289  & 966   & 2369  & 678   & 2235 \\
    {\color{black}AdvSKM} & {\color{black}13313} & {\color{black}5012}  & {\color{black}18444} & {\color{black}2344}  & {\color{black}4406}  & {\color{black}14624} & {\color{black}6601}  & {\color{black}4516}  & {\color{black}11857} & {\color{black}6332}  & {\color{black}4474}  & {\color{black}5218}  & {\color{black}8095} \\
    {\color{black}CoDATs} & {\color{black}1043}  & {\color{black}2854}  & {\color{black}5074}  & {\color{black}836}   & {\color{black}\underline{1533}}  & {\color{black}5094}  & {\color{black}1300}  & {\color{black}3993}  & {\color{black}2352}  & {\color{black}1065}  & {\color{black}2205}  & {\color{black}697}   & {\color{black}2337} \\
    CtsADA & 1467  & 1965  & 6832  & 4136  & 29153 & 22515 & 47682 & 17418 & \underline{2138}  & 13019 & 15436 & 1593  & 13613 \\
    CLUDA & 2686  & 4754  & 6337  & 873   & 3501  & 7159  & 2389  & 3232  & 3078  & 1658  & 5566  & 1093  & 3527 \\
        \midrule
    SEA   & \underline{922}  & \underline{1302}  & 3786  & \textbf{287} & {2090}  & \underline{3782}  & \underline{1163}  & \textbf{2990} & {2278}  & \underline{939}   & \textbf{1930} & \underline{635}   & \underline{1842} \\
    SEA++  & \textbf{881} & \textbf{803} & \textbf{2669} & \underline{295}   & \textbf{1510} & \textbf{3220} & \textbf{1117} & \underline{3045}  & \textbf{1558} & \textbf{859} & \underline{2072}  & \textbf{572} & \textbf{1550} \\
    \bottomrule
    \bottomrule
    \end{tabular}%
  \label{tab:SOTARUL}%
\end{table*}%
\begin{table*}[htbp]
  \centering
  \caption{The Comparisons with SOTAs in Opportunity HAR}
    \begin{tabular}{lccccccccccccc}
    \toprule
    \toprule
    Models & \textbf{1$\to$2} & \textbf{1$\to$3} & \textbf{1$\to$4} & \textbf{2$\to$1} & \textbf{2$\to$3} & \textbf{2$\to$4} & \textbf{3$\to$1} & \textbf{3$\to$2} & \textbf{3$\to$4} & \textbf{4$\to$1} & \textbf{4$\to$2} & \textbf{4$\to$3} & \textbf{Avg.} \\
    \midrule
    {\color{black}Target} & {\color{black}91.90} & {\color{black}97.95} & {\color{black}95.60} & {\color{black}98.40} & {\color{black}97.95} & {\color{black}95.60} & {\color{black}98.40} & {\color{black}91.90} & {\color{black}95.60} & {\color{black}98.40} & {\color{black}91.90} & {\color{black}97.95} & {\color{black}95.96} \\
    Source & 47.50 & 64.50 & 60.83 & 73.00 & 71.00 & 65.67 & 64.83 & 56.67 & 72.17 & 66.67 & 65.50 & 74.67 & 65.25 \\
    \midrule
    DDC   & 40.50 & 58.50 & 55.50 & 43.35 & 58.30 & 55.50 & 45.00 & 40.50 & 55.50 & 45.00 & 40.50 & 58.50 & 49.72 \\
    Coral & 78.55 & 82.71 & 78.17 & 81.58 & 79.93 & 79.07 & 78.50 & 75.40 & 84.88 & 78.56 & 80.38 & 87.88 & 80.47 \\
    DANN  & 69.62 & 85.12 & 79.00 & 85.62 & 81.62 & 74.88 & 76.50 & 75.75 & 86.12 & 84.25 & 74.12 & 86.62 & 79.94 \\
    BNM   & 44.00 & 61.00 & 52.00 & 45.00 & 61.00 & 52.00 & 45.00 & 44.00 & 52.00 & 45.50 & 44.00 & 61.00 & 50.54 \\
    LMMD  & 56.25 & 61.17 & 59.08 & 45.33 & 58.50 & 55.50 & 49.42 & 46.83 & 57.33 & 59.00 & 44.75 & 61.75 & 54.58 \\
    SDAT  & 84.92 & 72.42 & 77.00 & 87.25 & 83.00 & 76.17 & 87.75 & \textbf{84.25} & 87.25 & 74.58 & 66.75 & 91.50 & 81.07 \\
    {\color{black}AdvSKM} & {\color{black}58.10} &{\color{black} 58.25} & {\color{black}51.75} & {\color{black}45.40} & {\color{black}40.95} & {\color{black}48.40} & {\color{black}69.60} & {\color{black}72.20} & {\color{black}70.60} & {\color{black}68.75} & {\color{black}55.50} & {\color{black}58.20} & {\color{black}58.14} \\
    {\color{black}CoDATs} & {\color{black}80.17} & {\color{black}83.33} & {\color{black}82.33} & {\color{black}\underline{88.67}} & {\color{black}80.50}  & {\color{black}81.00}    & {\color{black}\underline{89.25}} & {\color{black}79.12} & {\color{black}85.75} & {\color{black}84.67} & {\color{black}79.17} & {\color{black}86.17} & {\color{black}83.34} \\
    CtsADA & 80.00 & 73.67 & 76.08 & 69.75 & 78.17 & 79.08 & 67.75 & 77.33 & 64.75 & 74.58 & 52.17 & 75.00 & 72.36 \\
    SLARDA & 80.38 & 80.50 & 79.50 & 82.75 & 78.88 & 79.25 & 75.88 & 78.62 & 70.12 & 80.62 & 84.00 & 77.25 & 78.98 \\
    CLUDA & 70.45 & 73.75 & 74.40 & 73.20 & 74.25 & 73.95 & 61.65 & 55.10 & 64.30 & 65.75 & 61.05 & 71.55 & 68.28 \\
        \midrule
    SEA   & \underline{86.35} & \underline{86.15} & \underline{86.00} & 86.35 & \textbf{84.35} & \underline{81.75} & 87.25 & 83.05 & \textbf{88.70} & \underline{84.90} & \underline{84.25} & \underline{92.25} & \underline{85.42} \\
    SEA++  & \textbf{86.85} & \textbf{90.30} & \textbf{90.85} & \textbf{90.25} & \underline{83.85} & \textbf{85.40} & \textbf{89.80} & \underline{83.10} & \underline{88.10} & \textbf{91.70} & \textbf{86.95} & \textbf{92.55} & \textbf{88.31} \\
    \bottomrule
    \bottomrule
    \end{tabular}%
  \label{tab:SOTAHAR}%
\end{table*}%

We compared SEA and SEA++ with State-Of-The-Art methods (SOTAs), including general UDA methods and TS UDA methods. General UDA methods were designed to be applicable to various tasks and domains, regardless of the specific data type, including DDC \cite{https://doi.org/10.48550/arxiv.1412.3474}, Deep Coral \cite{Sun2017}, DANN \cite{Ganin2017}, BNM \cite{Cui_2020_CVPR}, LMMD \cite{9085896}, and SDAT \cite{rangwani2022closer}. TS UDA methods were specifically designed for TS data, including AdvSKM \cite{liu2021adversarial}, CoDATs \cite{wilson2020multi}, CtsADA \cite{9234721}, SLARDA \cite{9804766}, and CLUDA \cite{vayyat2022cluda}. For fair comparisons, all methods were run ten times based on the same feature extractor (i.e., GNN-LSTM). LMMD and SLARDA were designed specifically for multi-class classification UDA, so they only had results on Opportunity HAR. Notably, in the previous SEA paper, different cross-domain scenarios were based on various sample lengths. For example, the cross-domain scenario 1$\to$2 was based on a time length of 72 while 3$\to$4 was based on a time length of 45. However, this training setting may be inapplicable in real-world systems where obtaining samples with a fixed time length is desired. Thus, all results in this paper, particularly in the context of the C-MAPSS dataset, have been re-conducted based on a consistent time length of 60 timestamps.
Additionally, we have included the results of Target-only (Target) and Source-only (Source) baselines for comparison. The target-only baseline represents the upper bound of UDA, where training and testing were performed on the target domain only. The source-only baseline represents the lower bound of UDA, where only the source data were used for training without any UDA methods applied.

TABLE \ref{tab:SOTARUL} shows the RMSE and Score results for 12 cross-domain scenarios in C-MAPSS. From the results, we observe that both SEA and SEA++ achieve better performance than SOTAs in most cross-domain scenarios. Specifically, compared to the methods excluding SEA++, SEA performs the best performance in 9 out of 12 scenarios and the second-best performance in the remaining scenarios with very marginal gaps compared to the best methods. We utilize the RMSE results in C-MAPSS for illustration. In the cross-domain scenario 1$\to$3, SEA outperforms the second-best method by a significant 9.6\%. In scenario 1$\to$4, SEA is the second-best method and is only 0.6\% weaker than the top-performing method. These results demonstrate the effectiveness of SEA compared to SOTAs. However, SEA exhibits minor limitations that make it slightly less effective in specific cases, motivating the development of SEA++. 

Regarding the results of SEA++, it not only achieves better performance in scenarios where SEA performs the best but also achieves the best performance in 2 out of 3 scenarios where SEA is the second-best. For example, in the cross-domain scenario 1$\to$4 where SEA is not the top-performing, SEA++ improves by 4.1\% compared to the best conventional method. Although SEA++ is slightly weak in 2$\to$3, the gap is further reduced, being only 0.3\% weaker than the best method. 
In terms of average RMSE results, SEA outperforms the best conventional methods by 9.2\%. With improved endo-feature alignment and multi-graph alignment, SEA++ further enhances the performance, surpassing SEA by 2.45\% and outperforming the best conventional method by 11.4\%. 
Similar improvements can be observed in the Score results of C-MAPSS and Opportunity HAR in TABLE \ref{tab:SOTAHAR}. Taking the average accuracy in Opportunity HAR for illustration, SEA improves by 2.08\% compared to the best conventional method. Additionally, SEA++ outperforms SEA by 2.89\% and outperforms the best conventional method by 4.97\%. 

These results indicate the effectiveness of SEA, showing the necessity of aligning sensor-level information. Furthermore, SEA++ addresses the minor limitations of SEA, resulting in improved performance that establishes its state-of-the-art status compared to traditional methods.


\subsection{Ablation Study}

\begin{table*}[htbp]
  \centering
  {\color{black}
  \caption{The Ablation Study in C-MAPSS (\textbf{R}: RMSE; \textbf{S}: Score)}
  \begin{threeparttable}
    \begin{tabular}{lccccccccccccc}
    \toprule
    \toprule
    Variants (\textbf{R}) & \textbf{1$\to$2} & \textbf{1$\to$3} & \textbf{1$\to$4} & \textbf{2$\to$1} & \textbf{2$\to$3} & \textbf{2$\to$4} & \textbf{3$\to$1} & \textbf{3$\to$2} & \textbf{3$\to$4} & \textbf{4$\to$1} & \textbf{4$\to$2} & \textbf{4$\to$3} & \textbf{Avg.} \\
    \midrule
    w/ Exo (w/o endo) & 16.96 & 27.74 & 28.28 & 14.47 & 33.70 & 28.18 & 32.88 & 25.45 & 19.98 & 30.75 & 23.19 & 20.03 & 25.13 \\
    \midrule
    w/ Exo + SCA & 15.96 & 23.83 & 25.40 & 14.12 & 27.32 & 27.13 & 26.89 & 23.97 & 19.87 & 25.94 & 20.62 & 17.97 & 22.42 \\
    w/ Exo + SCA + MGA & 15.96 & 21.95 & 22.68 & 13.65 & 28.93 & 26.31 & 27.98 & 21.64 & 18.70 & 26.15 & 20.31 & 17.33 & 21.80 \\
    w/ Exo + SCA + 2-order & 15.94 & 21.61 & 23.05 & 13.32 & 28.76 & 26.94 & 23.60 & 20.73 & 18.41 & 22.40 & 19.85 & 17.25 & 20.99 \\
    w/ Exo + iSCA & 15.90 & 21.39 & 22.62 & 13.28 & 27.91 & 26.08 & 22.01 & 19.86 & 18.30 & 22.37 & 19.55 & 17.18 & 20.54 \\
    \midrule
    w/ Exo + SFA & 16.71 & 22.03 & 24.96 & 14.03 & 29.01 & 26.15 & 26.71 & 21.28 & 19.71 & 25.97 & 19.63 & 17.30 & 21.96 \\
    w/ Exo + SFA + MGA & 16.11 & 21.61 & 22.47 & 13.75 & 28.97 & 26.05 & 27.35 & 21.20 & 18.97 & 23.67 & 19.58 & 17.08 & 21.40 \\
    w/ Exo + SFA + 2-order & 16.15 & 22.08 & 22.57 & 13.77 & 28.51 & 26.45 & 23.99 & 20.67 & 18.19 & 22.57 & 19.87 & 17.85 & 21.06 \\
    w/ Exo + iSFA & 15.90 & 20.37 & 22.47 & 13.59 & 27.80 & 25.34 & 21.44 & 19.95 & 18.10 & 20.54 & 19.30 & 17.28 & 20.17 \\
    \midrule
    SEA++ & 15.41 & 19.38 & 22.12 & 13.00 & 24.60 & 24.90 & 19.19 & 19.25 & 17.75 & 18.43 & 18.16 & 16.79 & 19.08 \\
    \midrule
    \midrule
    Variants (\textbf{S}) & \textbf{1$\to$2} & \textbf{1$\to$3} & \textbf{1$\to$4} & \textbf{2$\to$1} & \textbf{2$\to$3} & \textbf{2$\to$4} & \textbf{3$\to$1} & \textbf{3$\to$2} & \textbf{3$\to$4} & \textbf{4$\to$1} & \textbf{4$\to$2} & \textbf{4$\to$3} & \textbf{Avg.} \\
    \midrule
    w/ Exo (w/o endo) & 1156  & 1863  & 7246  & 406   & 3705  & 5160  & 17374 & 17854 & 2240  & 10247 & 6635  & 1224  & 6259 \\
    \midrule
    w/ Exo + SCA & 1101  & 1537  & 6392  & 360   & 2358  & 4657  & 11756 & 7351  & 2458  & 4082  & 3925  & 735   & 3893 \\
    w/ Exo + SCA + MGA & 980   & 979   & 2936  & 356   & 2255  & 3893  & 9340  & 7284  & 1974  & 4052  & 3652  & 664   & 3197 \\
    w/ Exo + SCA + 2-order & 995   & 1070  & 2838  & 328   & 2517  & 4287  & 3489  & 5954  & 1820  & 1976  & 3207  & 696   & 2431 \\
    w/ Exo + iSCA & 978   & 1045  & 2674  & 306   & 2076  & 3704  & 2270  & 3906  & 1571  & 1815  & 3084  & 679   & 2009 \\
    \midrule
    w/ Exo + SFA & 1052  & 1445  & 4347  & 368   & 2152  & 3855  & 6638  & 7266  & 2111  & 4406  & 3082  & 654   & 3115 \\
    w/ Exo + SFA + MGA & 949   & 1194  & 2634  & 364   & 1873  & 3697  & 6193  & 6023  & 1915  & 2443  & 3048  & 627   & 2580 \\
    w/ Exo + SFA + 2-order & 987   & 1099  & 2674  & 367   & 2233  & 4214  & 3654  & 7640  & 1621  & 2212  & 3208  & 707   & 2551 \\
    w/ Exo + iSFA & 946   & 1083  & 2634  & 342   & 1873  & 3598  & 1902  & 4515  & 1694  & 1395  & 2814  & 630   & 1952 \\
    \midrule
    SEA++ & 881   & 803   & 2669  & 295   & 1510  & 3220  & 1117  & 3045  & 1558  & 859   & 2072  & 572   & 1550 \\
    \bottomrule
    \bottomrule
    \end{tabular}%
    \begin{tablenotes}
        \footnotesize
        \item **MGA represents Multi-Graph Alignment; 2-order represents high-order statistical alignment.
    \end{tablenotes}
  \label{tab:ablarul}%
\end{threeparttable}
}
\end{table*}%

\begin{table*}[htbp]
  \centering
  {\color{black}
  \caption{The Ablation Study in Opportunity HAR}
    \begin{tabular}{lccccccccccccc}
    \toprule
    \toprule
    Variants & \textbf{1$\to$2} & \textbf{1$\to$3} & \textbf{1$\to$4} & \textbf{2$\to$1} & \textbf{2$\to$3} & \textbf{2$\to$4} & \textbf{3$\to$1} & \textbf{3$\to$2} & \textbf{3$\to$4} & \textbf{4$\to$1} & \textbf{4$\to$2} & \textbf{4$\to$3} & \textbf{Avg.} \\
    \midrule
    w/ Exo (w/o endo) & 78.55 & 82.71 & 78.17 & 81.58 & 79.93 & 79.07 & 78.50 & 75.40 & 84.88 & 78.56 & 80.38 & 87.88 & 80.47 \\
    \midrule
    w/ Exo + SCA & 84.45 & 84.90 & 82.20 & 84.00 & 81.95 & 81.00 & 81.50 & 75.81 & 86.95 & 81.80 & 81.60 & 87.35 & 82.79 \\
    w/ Exo + SCA + MGA & 85.00 & 86.75 & 84.05 & 88.25 & 82.60 & 82.20 & 83.25 & 79.75 & 85.80 & 83.90 & 81.85 & 88.35 & 84.31 \\
    w/ Exo + SCA + 2-order & 85.20 & 87.40 & 84.90 & 88.35 & 82.30 & 82.00 & 81.10 & 76.75 & 87.15 & 81.35 & 84.40 & 87.30 & 84.02 \\
    w/ Exo + iSCA & 85.70 & 87.20 & 88.20 & 89.10 & 81.30 & 84.10 & 86.75 & 79.15 & 88.05 & 89.80 & 85.50 & 87.50 & 86.03 \\
    \midrule
    w/ Exo + SFA & 82.94 & 83.62 & 80.38 & 84.25 & 83.45 & 80.55 & 78.80 & 75.38 & 86.20 & 79.05 & 81.90 & 87.50 & 82.00 \\
    w/ Exo + SFA + MGA & 85.95 & 87.60 & 85.75 & 87.05 & 82.75 & 82.90 & 79.60 & 79.05 & 87.35 & 83.50 & 79.95 & 88.90 & 84.20 \\
    w/ Exo + SFA + 2-order & 84.50 & 87.40 & 84.25 & 87.00 & 84.90 & 82.45 & 82.15 & 80.95 & 87.75 & 85.45 & 83.60 & 88.45 & 84.90 \\
    w/ Exo + iSFA & 85.70 & 87.90 & 85.85 & 87.90 & 83.70 & 82.90 & 86.45 & 82.05 & 86.90 & 87.10 & 84.75 & 90.15 & 85.95 \\
    \midrule
    SEA++  & 86.85 & 90.30 & 90.85 & 90.25 & 83.85 & 85.40 & 89.80 & 81.50 & 88.10 & 91.70 & 86.95 & 92.55 & 88.18 \\
    \bottomrule
    \bottomrule
    \end{tabular}%
  \label{tab:ablahar}
  }%
\end{table*}%


To comprehensively evaluate the effectiveness of our endo-feature alignment and the associated improvements, we have conducted extensive ablation studies consisting of nine different variants.
These variants can be categorized into three aspects.
The first aspect compares our method with a baseline that only performs exo-feature alignment without utilizing any sensor-level alignment.
In the second and third aspects, we consider variants that combine exo-feature alignment with one of the endo-feature alignment techniques: SCA or SFA. Each aspect further includes four variants based on different improvements.
We utilize the variants with SCA for illustration. The ``w/ Exo + SCA" variant represents the use of the vanilla SCA for sensor correlation alignment. By introducing Multi-Graph Alignment (MGA), we have the ``w/ Exo + SCA + MGA" variant, while still using alignment with first-order statistics.
By introducing alignment with second-order statistics without MGA, we have the variant ``w/ Exo + SCA + 2-order".
Finally, by incorporating both improvements, i.e., MGA and second-order statistic alignment, we obtain the ``w/ Exo + iSCA" variant.
The same variations were applied for SFA, resulting in four additional variants.

TABLE \ref{tab:ablarul} and TABLE \ref{tab:ablahar} show the ablation study on C-MAPSS and Opportunity HAR respectively. Focusing on the average RMSE results on C-MAPSS for illustration, we consider the variant ``w/ Exo" as the baseline. 
We can first observe that introducing SCA and SFA leads to improvements of 10.8\% and 12.6\%, respectively, which indicates the effectiveness of endo-feature alignment. 
As the endo-feature alignment still has minor limitations, we further introduce MGA and second-order statistic alignment to improve the sensor-level alignment. 
Introducing MGA leads to performance improvements of 13.3\% (SCA) and 14.9\% (SFA), while introducing second-order statistic alignment improves it by 16.5\% (SCA) and 16.2\% (SFA). The improvements indicate that MGA and second-order statistic alignment both show effectiveness for better sensor-level alignment. Thus, it is useful to combine them to further improve the performance of our method. Combining both parts, the variants ``w/ iSCA" and ``w/ iSFA" improve the performance by 18.3\% and 19.7\% respectively compared to the baseline. Finally, both the sensor features and sensor correlations can be aligned simultaneously. Thus, by combining iSCA and iSFA for endo-feature alignment, we achieve a remarkable improvement of 24.1\% compared to the baseline.
Similar trends are also observed in the experiments conducted on Opportunity HAR, as shown in TABLE \ref{tab:ablahar}. For example, the average accuracy of SEA++ increases by 7.61\% compared to the model with exo-feature alignment only. The results in the cross-domain scenarios of Opportunity HAR can further verify the effectiveness of our proposed modules.

These experiments highlight the significance of endo-feature alignment in achieving effective alignment at the sensor level for MTS-UDA. Furthermore, the additional modules for endo-feature alignment, namely MGA and high-order statistics alignment, prove to be highly effective in further improving our model's performance.

\subsection{Sensitivity Analysis}
Our method includes hyperparameters that require sensitivity analysis to evaluate their impact on the method's performance. We focus on four hyperparameters: $\lambda_{SCA}$, $\lambda_{SFA}$, number of heads, and patch sizes. The first two hyperparameters control the balance between sensor correlation alignment and sensor feature alignment. The number of heads determines the number of graph branches used to model spatial dependencies between sensors. The patch sizes determine the number of sequential graphs constructed. Conducting sensitivity analysis on these hyperparameters helps determine their sensitivity and find optimal values. {\color{black}Notably, the sensitivity analysis presented here pertains specifically to SEA++. Additional details regarding the analysis related to SEA can be found in our appendix.}

\subsubsection{Analysis for $\lambda_{SCA}$ and $\lambda_{SFA}$}
\begin{figure}[!ht]
    \centering
    \includegraphics[width = 1.\linewidth]{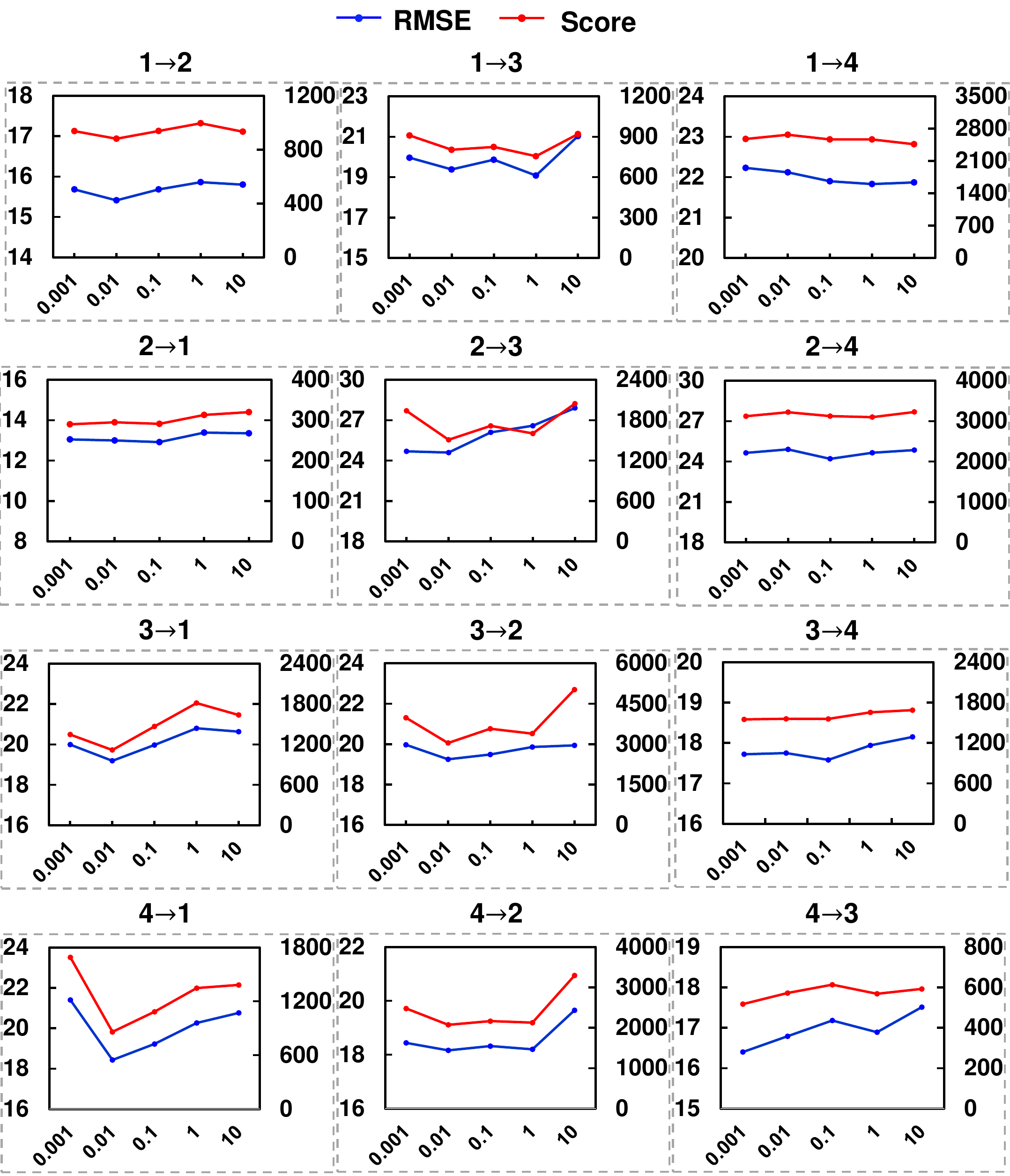}
    \caption{The sensitivity analysis for $\lambda_{SCA}$ on C-MAPSS.}
    \label{fig:HPRUL_full_SCA}
\end{figure}
\begin{figure}[!ht]
    \centering
    \includegraphics[width = 1.\linewidth]{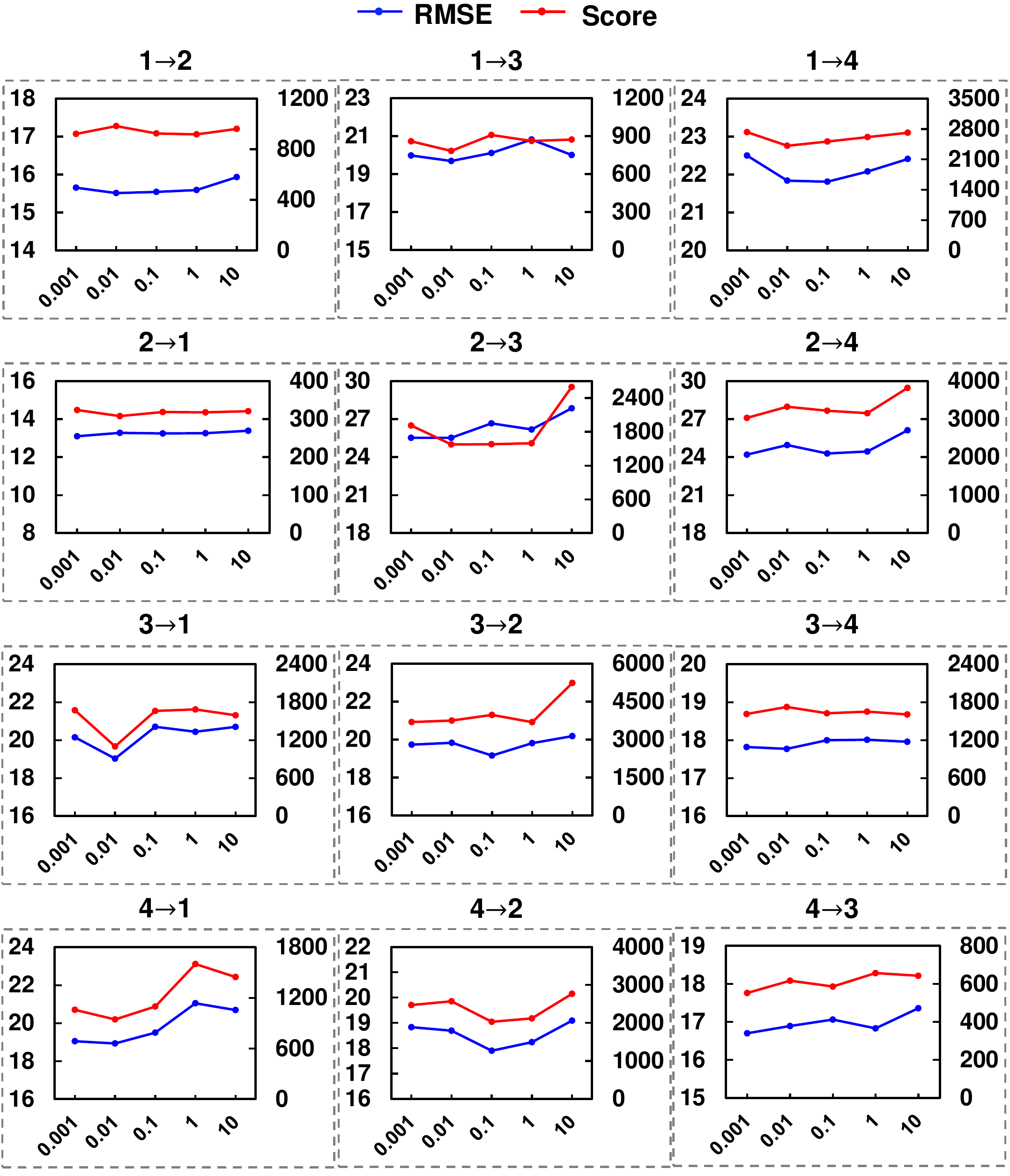}
    \caption{The sensitivity analysis for $\lambda_{SFA}$ on C-MAPSS.}
    \label{fig:HPRUL_full_SFA}
\end{figure}
\begin{figure}[!ht]
    \centering
    \includegraphics[width = 1.\linewidth]{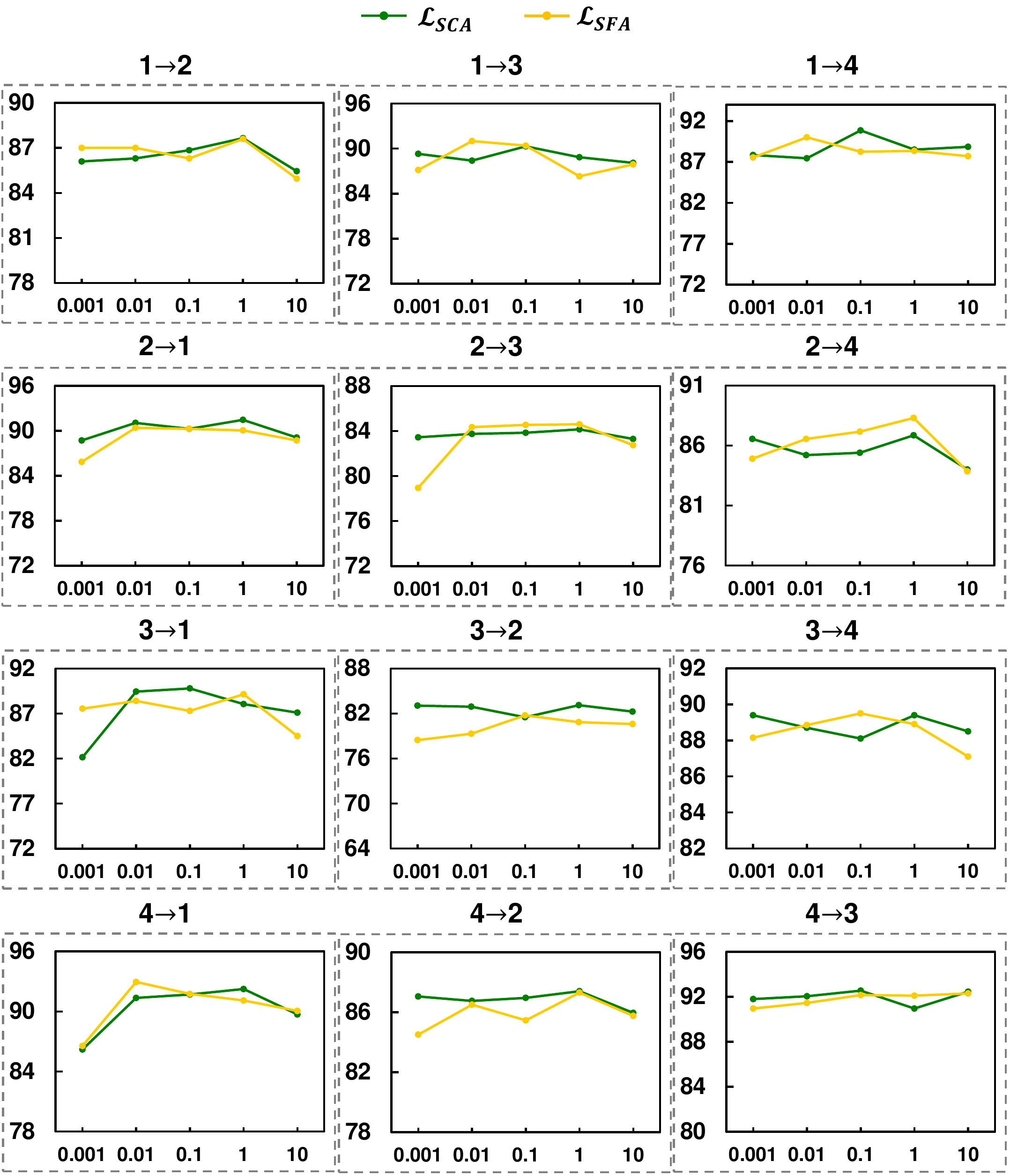}
    \caption{The sensitivity analysis for $\lambda_{SCA}$ and $\lambda_{SFA}$ on Opportunity HAR.}
    \label{fig:HPHAR_SCA_SFA}
\end{figure}
We have conducted experiments with various values of $\lambda_{SCA}$ and $\lambda_{SFA}$ varying from 0.001 to 10 with an interval of $10\times$. Fig. \ref{fig:HPRUL_full_SCA} and \ref{fig:HPRUL_full_SFA} present the analysis of $\lambda_{SCA}$ and $\lambda_{SFA}$ on C-MAPSS respectively, and Fig. \ref{fig:HPHAR_SCA_SFA} shows the analysis for $\lambda_{SCA}$ and $\lambda_{SFA}$ on Opportunity HAR. From the results, we can find similar trends for the two hyperparameters, which include two points. Firstly, our method exhibits insensitivity to changes in the hyperparameters in most cases. 
For example, our method is insensitive to $\lambda_{SCA}$ in cross-domain scenarios such as 1$\to$4, 2$\to$1, and 2$\to$4 of C-MAPSS, and 1$\to$3, 2$\to$3, and 4$\to$2 of Opportunity HAR. Similarly, regarding $\lambda_{SFA}$, we can also find the insensitivity in the scenarios 1$\to$2, 2$\to$1, and 3$\to$4 of C-MAPSS, and 1$\to$4, 3$\to$2, and 4$\to$3 of Opportunity HAR.
These results indicate that our method can achieve stable performance without requiring extensive hyperparameter tuning, making it applicable to real-world scenarios. Second, in the cases where our method is sensitive to the hyperparameters, our method tends to achieve good performance when their values are within the range of [0.001,1]. Taking the results on C-MAPSS as examples, we observe that our method achieves poor performance when $\lambda_{SCA}$ is set to 10 in the case of 3$\to$2, and when $\lambda_{SCA}$ is set to 0.001 in the case of 4$\to$1.
Similar trends can also be observed for $\lambda_{SFA}$, e.g., $\lambda_{SFA}$=10 in 2$\to$3. 
Based on these observations, we recommend setting the values of these hyperparameters within the range of 0.01 to 1, where our method consistently achieves good performance.

{\color{black}\subsubsection{Analysis for the number of heads}

\begin{figure}[!ht]
    \centering
    \includegraphics[width = 1.\linewidth]{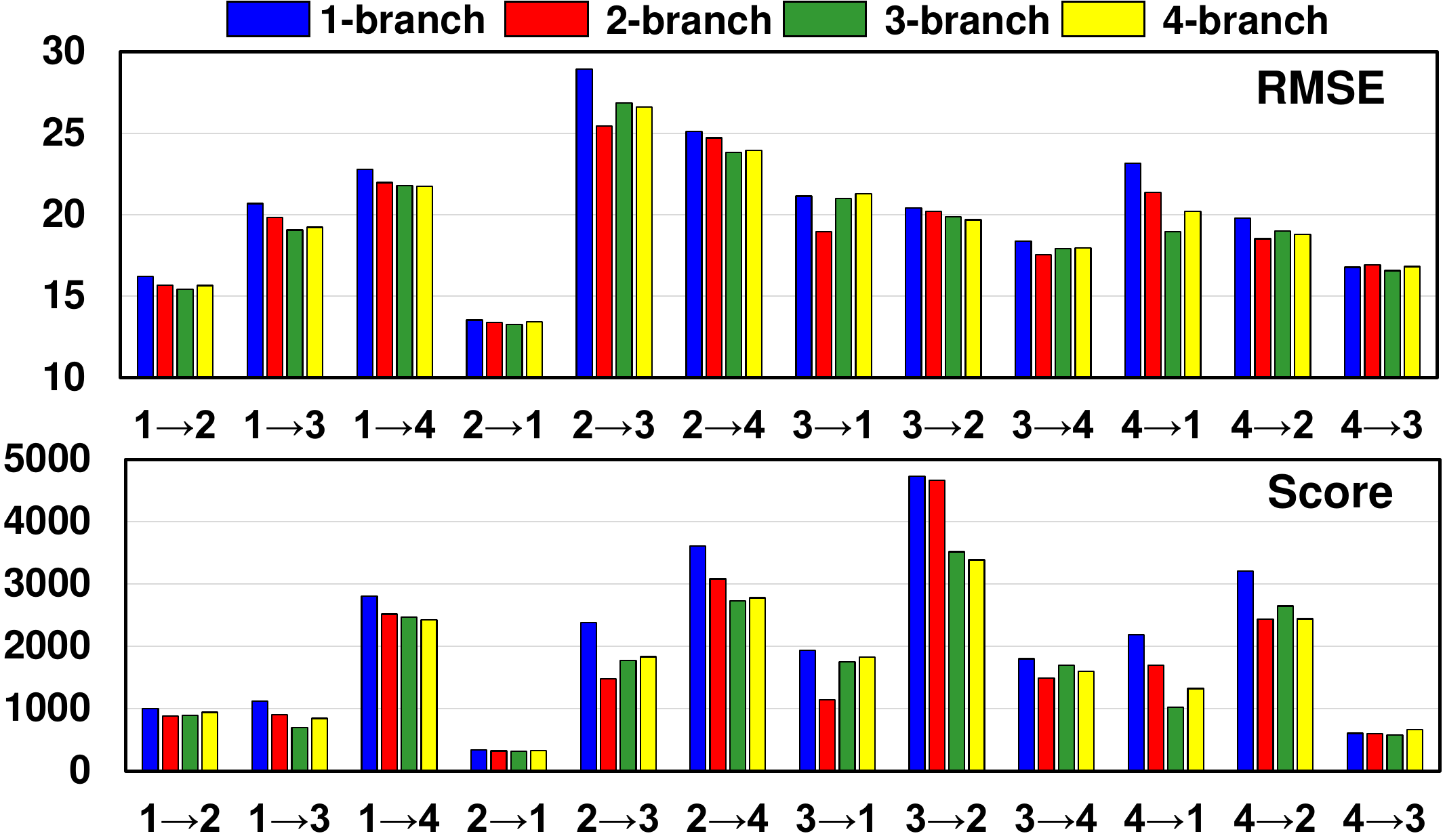}
    \caption{The sensitivity analysis for different heads on C-MAPSS.}
    \label{fig:HPRUL_full_heads}
\end{figure}
\begin{figure}[!ht]
    \centering
    \includegraphics[width = 1.\linewidth]{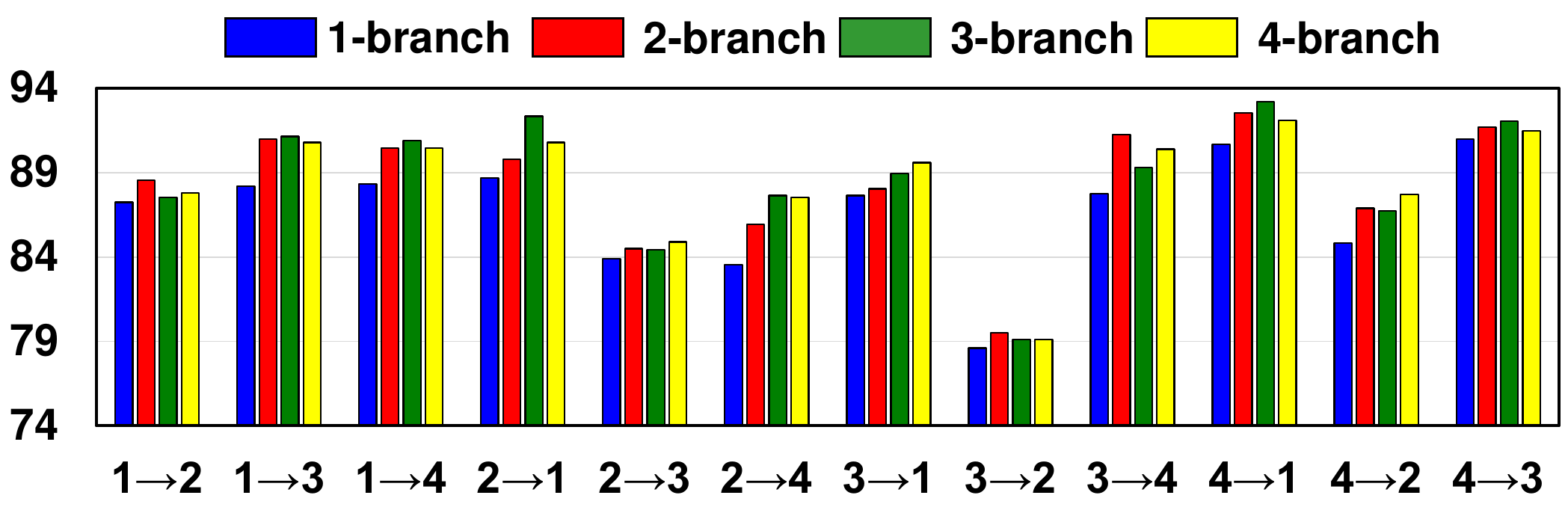}
    \caption{The sensitivity analysis for different heads on Opportunity HAR.}
    \label{fig:HPHAR_full_heads}
\end{figure}
For the construction of each graph with MSGC, we have multiple branches with differently initialized weights to model various aspects of sensor correlations. In this part, we aim to evaluate the effect of the number of branches on the model performance. Fig. \ref{fig:HPRUL_full_heads} and \ref{fig:HPHAR_full_heads} show the results on C-MAPSS and Opportunity HAR respectively. In our analysis, we analyze the effect using a maximum of 4 branches, as we observe that further increasing the number of branches does not yield significant improvements. The results consistently show that SEA++ obtains performance improvements with increasing the number of branches. Specifically, SEA++ with just one branch tends to exhibit poorer performance in most cases, such as in the 2$\to$3 cross-domain scenario of C-MAPSS and the 2$\to$4 scenario of Opportunity HAR. With increasing the number of branches, the performance improves. It is worth noting that the improvements become less pronounced when the number of branches is large enough, particularly when transitioning from 3 to 4 branches. This suggests that using 3 branches is generally sufficient for effectively capturing the spatial dependencies between sensors.
Overall, the analysis demonstrates the importance of leveraging multiple branches in MSGC for modeling the complex relationships among sensors. Increasing the number of branches allows our method to capture a more comprehensive understanding of sensor correlations, leading to improved performance for MTS-UDA.

\subsubsection{Analysis for patch size}
\begin{figure}[!ht]
    \centering
    \includegraphics[width = 1.\linewidth]{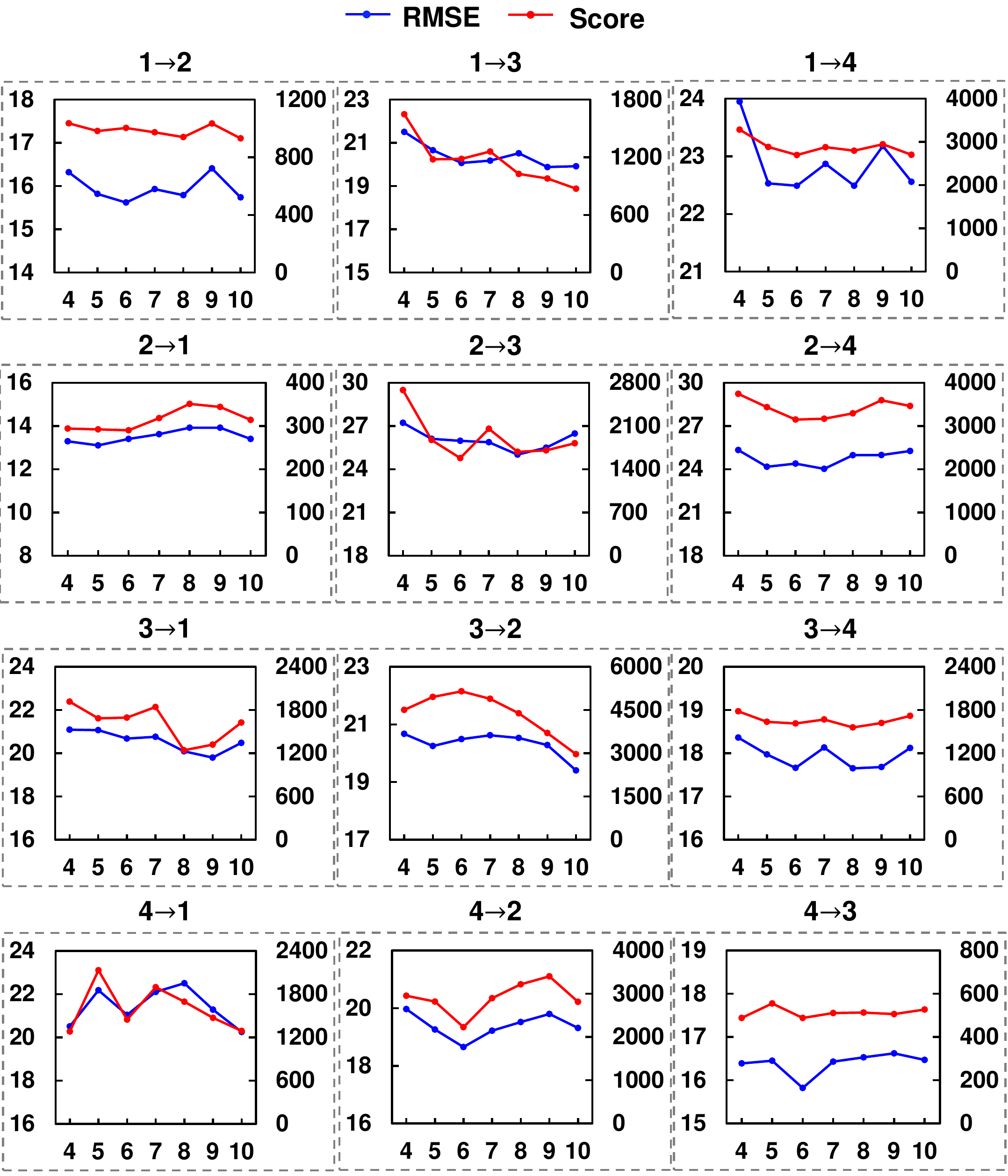}
    \caption{The sensitivity analysis for patch sizes on C-MAPSS.}
    \label{fig:HPRUL_full_windows}
\end{figure}
\begin{figure}[!ht]
    \centering
    \includegraphics[width = 1.\linewidth]{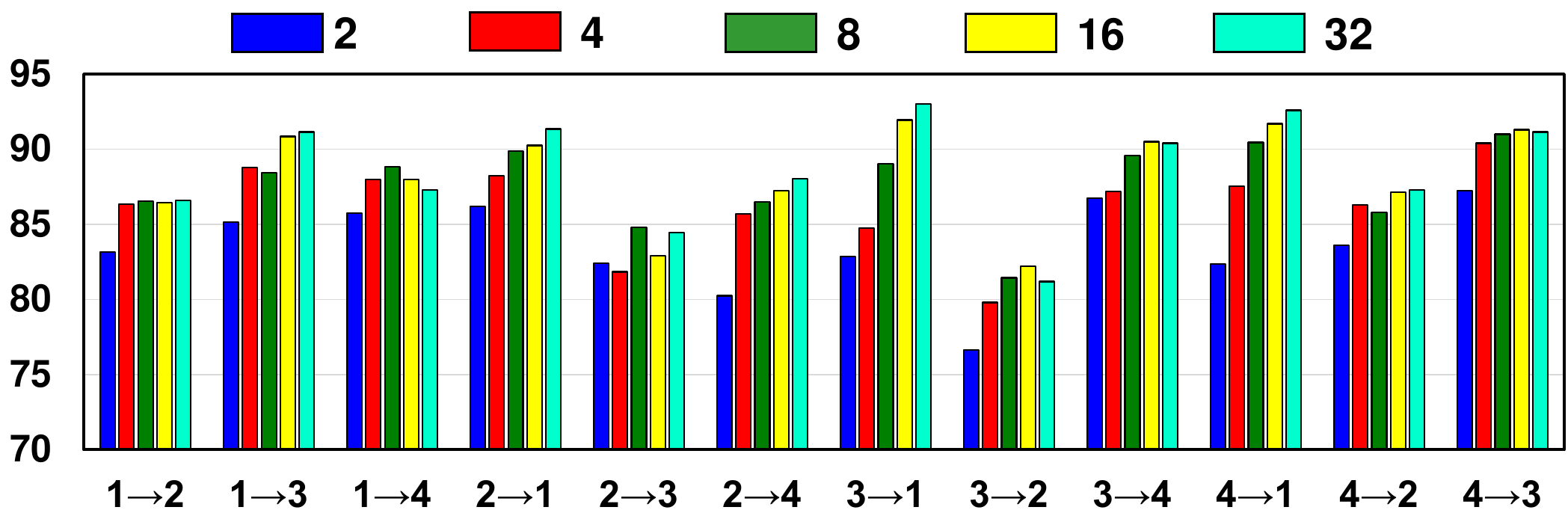}
    \caption{The sensitivity analysis for patch sizes on Opportunity HAR.}
    \label{fig:HPHAR_full_windows}
\end{figure}

To align the evolving distributions within MTS data, we segment an MTS sample into multiple patches with a size of $d$, which are then utilized to construct sequential graphs and conduct multi-graph alignment. Evaluating the effect of patch size $d$ is important, as the size affects the construction of sequential graphs. Notably, we conducted the analysis with different patch sizes for C-MAPSS and Opportunity HAR as their samples have different lengths. For C-MAPSS, the time length of each sample is fixed as 60, so we use the patch sizes within [4,5,6,7,8,9,10] for patch size analysis. For Opportunity HAR, the time length of each sample is fixed as 128, so we use the patch sizes within [2,4,8,16,32].

Based on the analysis of C-MAPSS in Fig. \ref{fig:HPRUL_full_windows} and Opportunity in Fig. \ref{fig:HPHAR_full_windows}, it is evident that relatively large patch sizes tend to result in better performance for our method. This observation holds true for several cross-scenarios, such as 3$\to$1, 3$\to$2, and 3$\to$4 of C-MAPSS, where the best performance is achieved with patch sizes of 8, 9, or 10. Similar observations can be found for Opportunity HAR, such as 1$\to$3, 2$\to$1, and 3$\to$1, where the best performance is achieved with patch sizes of 16 or 32.
Conversely, setting too small patch sizes leads to poor performance in most scenarios, such as 1$\to$3 and 1$\to$4 of C-MAPSS, and 1$\to$2 and 1$\to$3 of Opportunity HAR. 
The reason behind this trend lies in the construction of sequential graphs and the capturing of spatial dependencies for transferring across domains. By segmenting MTS data into patches, we construct sequential graphs whose spatial dependencies are captured for transferring across domains. In this process, too small patch sizes may not contain enough information to fully capture the complex spatial dependencies between sensors for each patch, resulting in suboptimal performance in transferring the dependency information. Therefore, it is recommended to use relatively large patch sizes to obtain better performance using SEA++ for MTS-UDA.}

\section{Conclusion}
\label{conclu}
In this paper, we formulate Multivariate Time-Series Unsupervised Domain Adaptation (MTS-UDA). We analyze the problems underlying this task and propose SEnsor Alignment (SEA) to address these issues. To reduce the domain discrepancy at both the local and global sensor levels, we design endo-feature alignment and exo-feature alignment. At the local sensor level, endo-feature alignment aligns sensor features and sensor correlations across domains, preventing misalignment at the sensor level. {\color{black}To enhance the endo-feature alignment, we further design SEA++, incorporating high-order statistic alignment and a multi-graph alignment technique. These enhancements are specifically designed to facilitate comprehensive alignment across domains and effectively capture the evolving data distributions within MTS data.} At the global sensor level, we enforce restrictions on global sensor features to reduce domain discrepancy. Our extensive experiments demonstrate the effectiveness of SEA and SEA++ for MTS-UDA. 


\bibliographystyle{IEEEtran}
\bibliography{references}
\end{document}